\documentclass[10pt,twocolumn,letterpaper]{article}

\usepackage[pagenumbers]{cvpr} 
\usepackage[accsupp]{axessibility}  

\usepackage{booktabs}       
\usepackage{amsfonts}       
\usepackage{nicefrac}       
\usepackage{microtype}      
\usepackage[table]{xcolor}
\usepackage{xcolor}         
\usepackage{graphicx} 
\usepackage{bm} 
\usepackage{amsmath}
\usepackage{multirow}
\usepackage{adjustbox}
\usepackage{algorithm} 
\usepackage{algpseudocode} 
\usepackage{subcaption}
\usepackage{makecell}
\usepackage{array} 
\usepackage{enumitem}
\usepackage{amssymb}
\usepackage{titletoc}

\definecolor{cvprblue}{rgb}{0.21,0.49,0.74}
\usepackage[pagebackref,breaklinks,colorlinks,allcolors=cvprblue]{hyperref}

\def\confName{CVPR}
\def\confYear{2026}

\title{Mind the Way You Select Negative Texts: \\ Pursuing the Distance Consistency in OOD Detection with VLMs}

\author{Zhikang Xu \textsuperscript{1,2}\hspace{1em}
Qianqian Xu \textsuperscript{3,5,}\thanks{Corresponding authors}\hspace{1em}
Zitai Wang \textsuperscript{3}\hspace{1em}
Cong Hua \textsuperscript{3,4}\hspace{1em} \\
Sicong Li \textsuperscript{1,2}\hspace{1em}
Zhiyong Yang \textsuperscript{4} \hspace{1em} 
Qingming Huang \textsuperscript{4,3,*} \\
{\textsuperscript{1}Institute of Information Engineering, Chinese Academy of Sciences} \\
{\textsuperscript{2}School of Cyber Security, University of Chinese Academy of Sciences} \\
{\textsuperscript{3}Institute of Computing Technology, Chinese Academy of Sciences} \\
{\textsuperscript{4}School of Computer Science and Technology, University of Chinese Academy of Sciences} \\
{\textsuperscript{5}Beijing Academy of Artificial Intelligence} \\
{\tt\small \{xuzhikang, lisicong\}@iie.ac.cn \hspace{2em} \{xuqianqian, wangzitai, huacong23z\}@ict.ac.cn} \\ 
{\tt\small \{yangzhiyong21, qmhuang\}@ucas.ac.cn }
}

\begin{document}
\maketitle
\begin{abstract}
  Out-of-distribution (OOD) detection seeks to identify samples from unknown classes, a critical capability for deploying machine learning models in open-world scenarios. Recent research has demonstrated that Vision-Language Models (VLMs) can effectively leverage their multi-modal representations for OOD detection. However, current methods often incorporate \textbf{intra-modal} distance during OOD detection, such as comparing negative texts with ID labels or comparing test images with image proxies. This design paradigm creates an inherent inconsistency against the \textbf{inter-modal} distance that CLIP-like VLMs are optimized for, potentially leading to suboptimal performance. To address this limitation, we propose InterNeg, a simple yet effective framework that systematically utilizes consistent inter-modal distance enhancement from textual and visual perspectives. From the textual perspective, we devise an inter-modal criterion for selecting negative texts. From the visual perspective, we dynamically identify high-confidence OOD images and invert them into the textual space, generating extra negative text embeddings guided by inter-modal distance. Extensive experiments across multiple benchmarks demonstrate the superiority of our approach. Notably, our InterNeg achieves state-of-the-art performance compared to existing works, with a 3.47\% reduction in FPR95 on the large-scale ImageNet benchmark and a 5.50\% improvement in AUROC on the challenging Near-OOD benchmark. Code will be available at \url{https://github.com/ZhikangXu0112/InterNeg}.
\end{abstract}    
\section{Introduction}
\definecolor{intra-modal}{RGB}{255,147,170}
\definecolor{inter-modal}{RGB}{122,180,168}
\begin{figure*}[t]
  \centering
  \includegraphics[width=\textwidth]{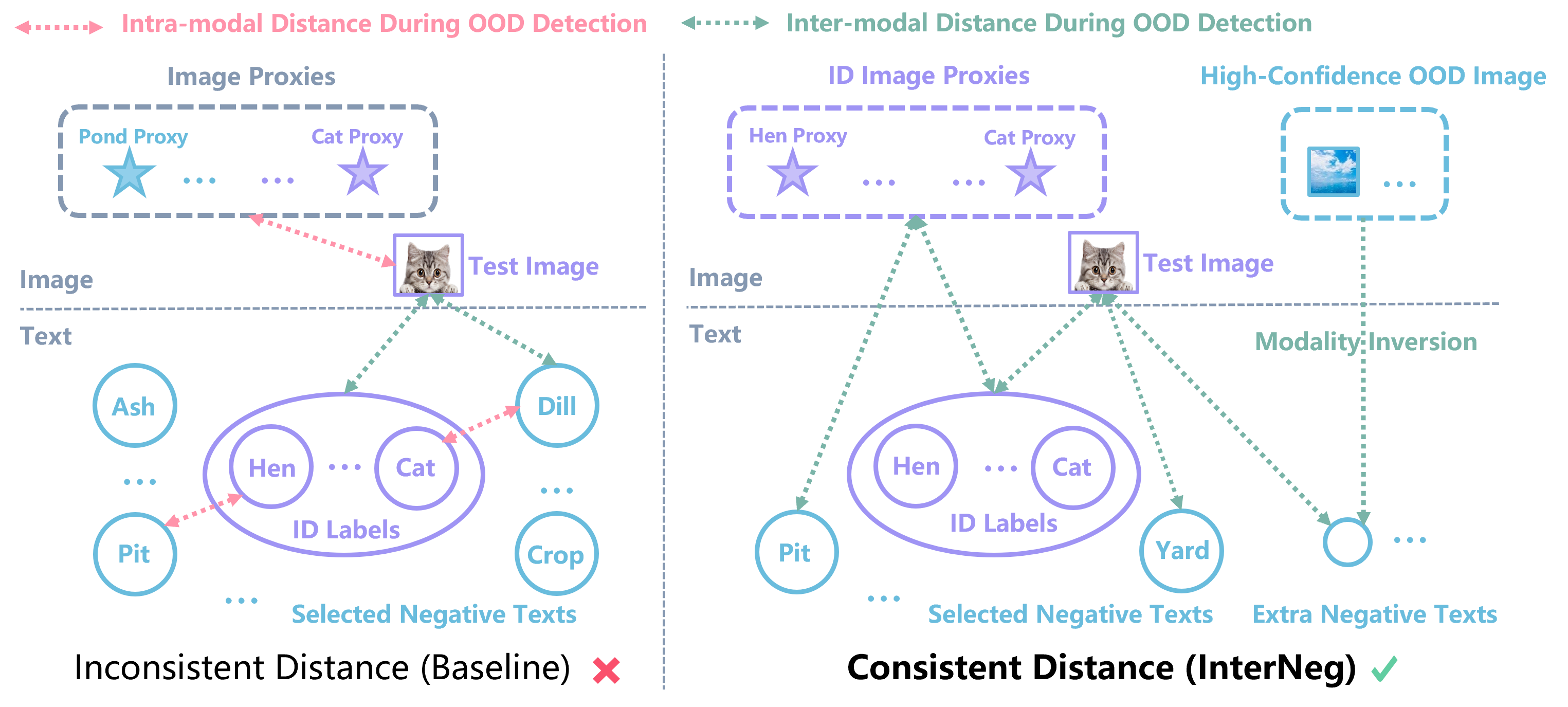}
    \caption{Comparison of Baseline and InterNeg. The baseline often incorporates \textcolor{intra-modal}{\textbf{intra-modal}} distance during OOD detection, which is inconsistent with the \textcolor{inter-modal}{\textbf{inter-modal}} distance that CLIP-like VLMs are optimized for. In contrast, InterNeg leverages consistent inter-modal distance during OOD detection, enhancing performance by inter-modal guided negative texts and extra negative text embeddings generated through modality inversion.}
  \label{fig::intro}
\end{figure*}
Most machine learning algorithms \cite{hua2024reconboost,zhang2025scrutinize,wang2026rethinking,ma2026learning,Ma_2025_CVPR} are designed based on the closed-world assumption, where training and testing datasets share the same label space. However, in real-world applications, models inevitably encounter samples from unknown classes, which are referred to as \emph{out-of-distribution} (OOD) data. In such scenarios, models often exhibit overconfidence, misclassifying OOD data into known classes \cite{DBLP:conf/cvpr/NguyenYC15}, posing significant risks in high-stakes scenarios like risk content identification, autonomous driving, and medical diagnosis \cite{DBLP:journals/corr/abs-2108-07258}. Consequently, OOD detection is crucial for building trustworthy and secure artificial intelligence systems and has garnered widespread attention \cite{DBLP:journals/tmlr/SalehiMHLRS22,DBLP:journals/ijcv/YangZLL24}.

Traditional visual OOD detection methods \cite{DBLP:conf/iclr/LiangLS18,DBLP:conf/nips/LiuWOL20,DBLP:conf/cvpr/LiCHY0J23,openauc} focus on training powerful representation models and designing OOD scoring functions based on a single image modality. Recently, inspired by the power of vision-language models (VLMs) such as CLIP \cite{DBLP:conf/icml/RadfordKHRGASAM21}, an increasing number of studies \cite{DBLP:conf/aaai/Esmaeilpour00022,DBLP:conf/nips/MingCGSL022,DBLP:conf/iclr/Jiang000LZ024,Fu_2025_WACV} have explored how to leverage the rich multi-modal knowledge to enhance OOD detection capabilities. Specifically, ZOC \cite{DBLP:conf/aaai/Esmaeilpour00022} and MCM \cite{DBLP:conf/nips/MingCGSL022} introduce zero-shot OOD detection, utilizing VLMs to identify OOD samples without the need for training on \emph{in-distribution} (ID) data. NegLabel \cite{DBLP:conf/iclr/Jiang000LZ024} proposes to mine massive negative texts that are semantically distant from ID labels in the textual space to further unlock the potential of VLMs in enhancing OOD detection performance. Owing to its simplicity and strong empirical performance, NegLabel has inspired a series of subsequent studies \cite{Fu_2025_WACV,DBLP:conf/nips/ZhangZ24,DBLP:conf/nips/Chen0X24} that build upon its framework to advance the field of OOD detection further. Remarkably, these methods even outperform those \cite{DBLP:conf/nips/MiyaiYIA23,DBLP:conf/iclr/NieZ0L0024,DBLP:conf/cvpr/BaiHCJHZ24,DBLP:conf/cvpr/LiP0MZ24,DBLP:journals/corr/abs-2502-00662} that require training on ID data.

However, as shown in Figure \ref{fig::intro}, existing methods \cite{DBLP:conf/iclr/Jiang000LZ024,DBLP:conf/nips/ZhangZ24} often adopt \textcolor{intra-modal}{\textbf{intra-modal}} (\textit{i.e.} image-image and text-text) distance during OOD detection, by either comparing negative texts with ID labels or comparing test images with image proxies. This strategy contradicts the \textcolor{inter-modal}{\textbf{inter-modal}} (\textit{i.e.} image-text) optimization objective of CLIP-style VLMs \cite{DBLP:conf/icml/RadfordKHRGASAM21,crossthegap}, potentially resulting in suboptimal performance. For instance, while some negative texts have a large intra-modal distance from ID labels, their inter-modal distance to the ID test image may still be smaller than that of the ID labels to the ID test image, potentially causing ID misclassification. In Section \ref{sec:misclassification}, we provide a detailed analysis to further explain this issue.

To address this limitation, we introduce InterNeg, a simple yet effective approach that explicitly utilizes consistent inter-modal distance from textual and visual perspectives without requiring training on ID or extra data. 
\textbf{From the textual perspective}, we propose an inter-modal guided negative text selection strategy by introducing an ID inter-modal base distance. 
\textbf{From the visual perspective}, we leverage high-confidence OOD image inversion to generate extra negative text embeddings during inference. Since high-confidence OOD images are determined by a fixed threshold, some noisy samples may be inadvertently included. To mitigate this issue, we also introduce an inter-modal guided dynamic filtering mechanism. Finally, after obtaining the inter-modal guided negative texts and extra negative text embeddings, our method effectively achieves enhanced performance.

Extensive experimental results demonstrate that our method achieves state-of-the-art performance on various OOD detection benchmarks. Specifically, on the large-scale ImageNet-1K traditional Four-OOD benchmark \cite{DBLP:conf/cvpr/HuangL21}, our approach yields a notable 3.47\%
reduction in FPR95 while improving AUROC by 0.77\% over the existing methods. Moreover, on the more challenging Near-OOD benchmark \cite{DBLP:conf/nips/YangWZZDPWCLSDZ22,DBLP:journals/corr/abs-2306-09301}, our method achieves a substantial 2.09\% reduction in FPR95 and a significant 5.50\% improvement in AUROC. In summary, our main contributions can be listed as follows:

\begin{itemize}
    \item To the best of our knowledge, we are the first to identify an inconsistency between the intra-modal distance during OOD detection and the inter-modal distance that CLIP-like VLMs are optimized for, which may lead to suboptimal performance.
    \item We propose InterNeg, a simple yet effective approach that leverages consistent inter-modal distance from both textual and visual perspectives without requiring training on any data. 
    \item Extensive experiments on multiple benchmarks demonstrate that our method achieves state-of-the-art performance. Moreover, ablation studies and further discussions also validate the effectiveness and robustness of our method.
\end{itemize}

\section{Related Work}
\label{sec:related}

\paragraph{Visual-based Traditional OOD Detection.} Traditional OOD detection methods only use single image modality, which can be broadly categorized into score-based \cite{DBLP:conf/iclr/HendrycksG17,DBLP:conf/cvpr/HuangL21,DBLP:conf/nips/LeeLLS18,DBLP:conf/iclr/LiangLS18,DBLP:conf/nips/LiuWOL20,DBLP:conf/nips/SunGL21,DBLP:conf/cvpr/Wang0F022}, density-based \cite{DBLP:conf/nips/RenLFSPDDL19,DBLP:conf/nips/XiaoYA20}, and distance-based \cite{DBLP:conf/icml/SunM0L22,DBLP:conf/nips/DuGML22}. Score-based methods leverage a well-trained classifier to derive better score functions. Density-based methods model the ID data probabilistically, flagging samples in low-density regions as OOD. Distance-based methods detect OOD samples by measuring feature space distances to ID prototypes. However, these methods are based solely on the visual modality, overlooking the potential benefits that integrating the textual modality could bring.

\paragraph{VLM-based OOD Detection.}
Recently, the advent of powerful pre-trained vision-language models (VLMs), such as CLIP \cite{DBLP:conf/icml/RadfordKHRGASAM21}, has inspired a growing number of research leveraging VLMs for OOD detection. These works can be broadly divided into few-shot and zero-shot learning approaches. LoCoOp \cite{DBLP:conf/nips/MiyaiYIA23} first proposes few-shot OOD detection, extracting regions from the local features of CLIP that are irrelevant to the ID data as regularization constraints for OOD detection. A subsequent series of works \cite{DBLP:conf/iclr/NieZ0L0024,DBLP:conf/cvpr/BaiHCJHZ24,DBLP:conf/cvpr/LiP0MZ24,DBLP:journals/corr/abs-2502-00662,DBLP:conf/icml/Hua00WBH25,lightfair} have focused on training negative text prompts to identify OOD samples. 

The zero-shot OOD detection task is first proposed by ZOC \cite{DBLP:conf/aaai/Esmaeilpour00022}, which trains an unknown-class text generator. However, this approach yields poor results when dealing with large-scale ID datasets such as ImageNet-1K. MCM \cite{DBLP:conf/nips/MingCGSL022} introduces a temperature-scaled maximum softmax probability as the OOD detection score, yet this method solely relies on ID labels without fully leveraging the open-world textual capabilities of VLMs. To enrich textual information, CLIPN \cite{DBLP:conf/iccv/WangLYL23} trained an additional negative text encoder using auxiliary datasets, while NegLabel \cite{DBLP:conf/iclr/Jiang000LZ024} selected numerous potential negative text labels from large lexical corpus (e.g., WordNet \cite{fellbaum1998wordnet}) to approximate OOD labels. EOE \cite{DBLP:conf/icml/CaoZZ0L024} designed multi-granularity prompts based on image similarity and employed large language models to generate potential OOD labels. Among these, NegLabel has been widely adopted in subsequent works \cite{Fu_2025_WACV,DBLP:conf/nips/Chen0X24,DBLP:conf/nips/ZhangZ24} due to its simplicity and efficiency. Building upon this foundation, CLIPScope \cite{Fu_2025_WACV} proposes a more robust OOD scoring method based on Bayesian theory, while CSP \cite{DBLP:conf/nips/Chen0X24} theoretically demonstrates that expanding the corpus size enhances OOD detection capability and proposes a practical expansion approach. Meanwhile, AdaNeg \cite{DBLP:conf/nips/ZhangZ24} employs a memory bank during inference to store high-confidence samples, effectively aligning with OOD label space by generating adaptive negative proxies.
\section{Preliminaries}


\paragraph{Problem Setups.}
Let $\mathcal{X}^{\text{ID}}$ and $\mathcal{Y}^{\text{ID}}=\{y_1,y_2,\cdots,y_C\}$ be the ID image space and ID label space, where $\mathcal{Y}$ consists text words like $\mathcal{Y}^{\text{ID}}=\{cat,dog,\cdots,goose\}$ and $C$ is the total number of ID classes. In closed-world scenarios, it is assumed that the training and testing data come from the same distribution $\mathcal{X}^{\text{ID}}$. However, in real-world applications, models may inevitably encounter samples from unknown classes, denoted by $\boldsymbol{x} \in \mathcal{X}^{\text{OOD}}$ and $y \in \mathcal{Y}^{\text{OOD}}$. In such scenarios, models may misclassify OOD data into ID classes with high confidence \cite{DBLP:conf/cvpr/NguyenYC15}. To address this issue, OOD detection \cite{DBLP:journals/ijcv/YangZLL24} is proposed to identify ID data and reject OOD data using a score function $S(\cdot)$ \cite{DBLP:conf/iclr/HendrycksG17,DBLP:conf/iclr/LiangLS18,DBLP:conf/nips/LiuWOL20,DBLP:conf/cvpr/LiCHY0J23}. Let $\mathcal{X}=\mathcal{X}^{\text{ID}}\cup {\mathcal{X}}^{\text{OOD}}$ and $\mathcal{X}^{\text{ID}}\cap {\mathcal{X}}^{\text{OOD}}=\varnothing $, given a test sample $\boldsymbol{x}\in \mathcal{X}$, the OOD detector $G(\cdot)$ can be defined as:
\begin{equation}
    G_{\gamma}(\boldsymbol{x}) = 
        \begin{cases} 
        \text{ID}, & \text{if } S(\boldsymbol{x}) \geq \gamma; \\
        \text{OOD}, & \text{otherwise,}
        \end{cases}
\end{equation}
where $\gamma \in \mathbb{R}$ is a predefined threshold to distinguish ID/OOD classes. The test sample $\boldsymbol{x}$ is detected as ID data if and only if $S(\boldsymbol{x}) \geq \gamma$.

\paragraph{CLIP-based VLMs.}
CLIP \cite{DBLP:conf/icml/RadfordKHRGASAM21} is a foundational VLM pre-trained on 400 million image-text pairs collected from the internet using self-supervised contrastive learning. It contains a text encoder $\mathcal{T}(\cdot)$ using the Transformer \cite{DBLP:conf/nips/VaswaniSPUJGKP17} architecture and an image encoder $\mathcal{I}(\cdot)$ using the ViT \cite{DBLP:conf/iclr/DosovitskiyB0WZ21} or ResNet \cite{DBLP:conf/cvpr/HeZRS16} architecture. Given a batch of $N$ image-text pairs $\{\boldsymbol{x}_i,y_i\}_{i=1}^N$, we extract the image features $\boldsymbol{h}_i \in \mathbb{R}^d$ and text features $\boldsymbol{e}_i \in \mathbb{R}^d$ as follows:
\begin{equation}
    \boldsymbol{h_i} = \mathcal{I}(\boldsymbol{x}_i),\quad \boldsymbol{e}_i=\mathcal{T}(\mathcal{E}(\text{prompt}(y_i))),\quad \forall i=1,2,\cdots,N,
\end{equation}
where $\text{prompt}(\cdot)$ represents the prompt template for input labels, \textit{e.g.}, \texttt{"a photo of [class]"}, $\mathcal{E}(\cdot)$ is word embedding function and $d$ is the embedding dimension. During training, CLIP optimizes a symmetric contrastive loss that explicitly minimizes the \textbf{inter-modal} distance (\textit{i.e.,} cosine similarity) between matched image-text pairs while maximizing it for unmatched pairs:
\begin{equation}
\begin{aligned}
\mathcal{L}_{\text{CLIP}} = -\frac{1}{N}\sum_{i=1}^N
&\left(
    \log\frac{e^{\cos(\boldsymbol{h}_i,\boldsymbol{e}_i)/\tau}}
           {\sum_{j=1}^N e^{\cos(\boldsymbol{h}_i,\boldsymbol{e}_j)/\tau}}
\right. \\
&+ \left.
    \log\frac{e^{\cos(\boldsymbol{e}_i,\boldsymbol{h}_i)/\tau}}
           {\sum_{j=1}^N e^{\cos(\boldsymbol{e}_i,\boldsymbol{h}_j)/\tau}}
\right),
\end{aligned}
\end{equation}
where $\text{cos}(\cdot,\cdot)$ denotes the cosine similarity and $\tau$ is the temperature parameter. During inference, for a image $\boldsymbol{x} \in \mathcal{X}^{\text{ID}},y\in\mathcal{Y}^{\text{ID}}$, the predictions are calculated by the cosine similarity between the image features $\boldsymbol{h}$ and text features $\boldsymbol{e}_i$:
\begin{equation}
    \hat{y} = \arg\max\limits_{y_i\in\mathcal{Y}^{\text{ID}}}\  \text{softmax}\{\text{cos}(\boldsymbol{h},\boldsymbol{e}_i)\}.
\end{equation}
The vanilla CLIP is proposed to perform zero-shot ID classification. Recently, it has been extended to zero-shot OOD detection.

\begin{figure*}[t]
  \centering
\includegraphics[width=\textwidth]{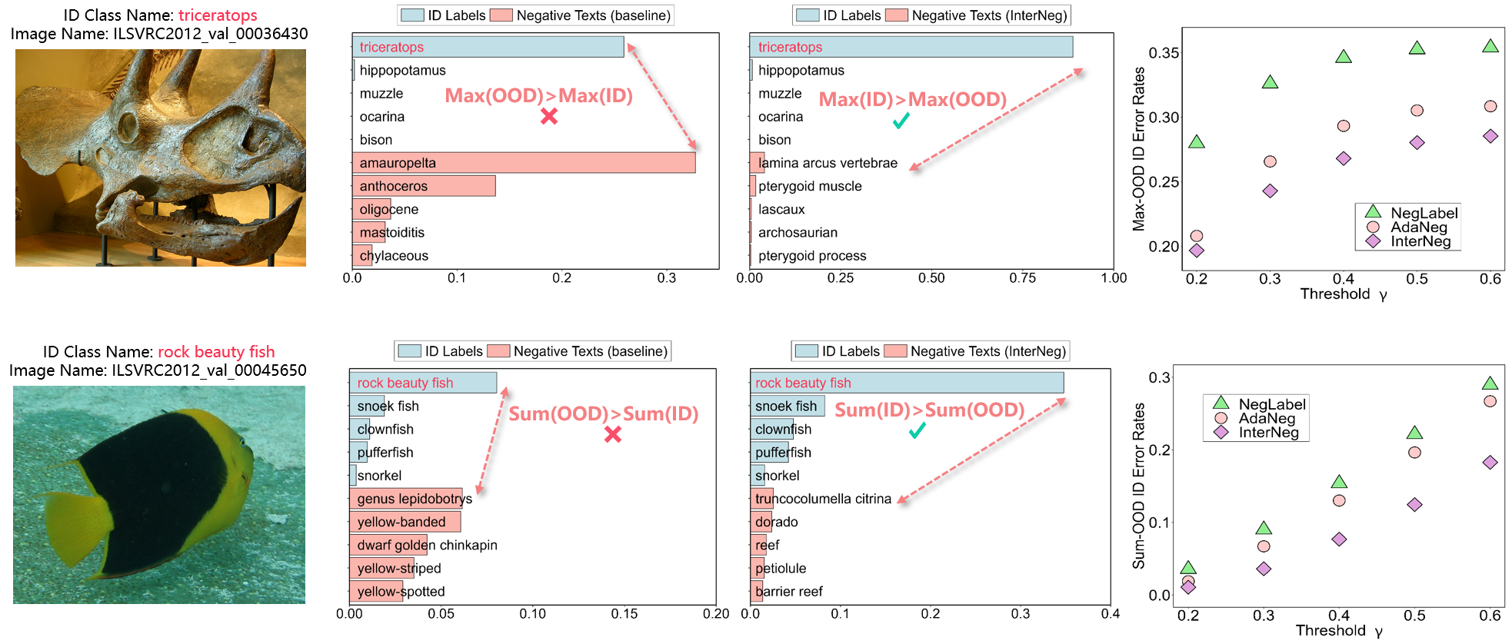}

  \caption{Two types of ID misclassification. \emph{First Row:} Max-OOD dominant ID misclassification. \emph{Second Row:} Sum-OOD dominant ID misclassification. \emph{Left:} Original ID image from ImageNet-1K with its class label and filename. \emph{Middle:} Top-5 softmax scores for ID labels and negative texts of baseline and our method. \emph{Right:} Max-OOD/Sum-OOD dominant ID error rates under different thresholds $\gamma$ of baseline and our method.}

  \label{fig::misclassification}
\end{figure*}

\paragraph{CLIP for OOD detection (NegLabel) \cite{DBLP:conf/iclr/Jiang000LZ024}.} As a straightforward and effective method, NegLabel has been widely used in recent works \cite{Fu_2025_WACV,DBLP:conf/nips/ZhangZ24,DBLP:conf/nips/Chen0X24}. Specifically, NegLabel proposes to mine a set of negative texts $\mathcal{Y}^-=\{y_{C+1},y_{C+2},\cdots,y_{C+M}\}$ from a large text corpus, where $\mathcal{Y}^- \cap \mathcal{Y}^{\text{ID}} = \varnothing$ and $M$ is the number of selected negative texts. These negative texts are selected based on intra-modal distance, which means they exhibit lower cosine similarity with ID labels, thereby serving as an approximation for the true OOD labels. Then, we can extend the text features as ID label features $\boldsymbol{e}_i$ and negative text features $\boldsymbol{e}_i^-\in \mathbb{R}^d$:
\begin{equation}
    \boldsymbol{e}_i^- = \mathcal{T}(\mathcal{E}(\text{prompt}(y_{i+C}))),\quad \forall i=1,2,\cdots,M,
\end{equation}
On top of this, NegLabel proposes an OOD score function to distinguish ID/OOD samples as follows:\

\begin{equation}
    S_{\text{NegLabel}}(\boldsymbol{x})=\frac{\sum_{i=1}^C e^{\text{cos}(\boldsymbol{h},\boldsymbol{e}_i)/\tau}}{\sum_{i=1}^C e^{\text{cos}(\boldsymbol{h},\boldsymbol{e}_i)/\tau}+\sum_{i=1}^M e^{\text{cos}(\boldsymbol{h},\boldsymbol{e}^-_i)/\tau}},
\end{equation}
where $\tau>0$ is the scaling temperature. In a word, NegLabel extends the label space to fully leverage the potential of utilizing VLMs for OOD detection.

\section{Methodology}

\subsection{Motivations}
\label{sec:misclassification}

Existing approaches, such as NegLabel and AdaNeg, utilize intra-modal distance during OOD detection. However, large intra-modal distances do not guarantee correspondingly large inter-modal distances, potentially leading to ID misclassification (\textit{i.e.}, misclassifying ID as OOD).
In Figure \ref{fig::misclassification}, we visualize the results of two ID images in the ImageNet-1K dataset and identify two distinct types of ID misclassification:
\begin{itemize}[itemsep=1pt, topsep=1pt, parsep=0pt, partopsep=0pt]
    \item \textbf{Max-OOD dominant} occurs when the highest-scoring negative (OOD) text surpasses the score of the highest-scoring in-distribution (ID) label, leading to ID misclassification. For instance, if the selected negative text "Amauropelta" receives a higher score than the ground-truth label "Triceratops," the ID input is misclassified.
    \item \textbf{Sum-OOD dominant} refers to cases where the total negative text score exceeds the total ID score, excluding cases already categorized as Max-OOD. For instance, the prior arts may select negative texts such as "genus lepidobotrys", "yellow-banded", "dwarf golden chinkapin", and so on. Although none of these individually dominate, their aggregated score surpasses that of the ground-truth label, ultimately leading to the misclassification of the ID input.
\end{itemize}

To quantify these phenomena, we compute the ID error rates for Max-OOD and Sum-OOD cases under varying threshold values $\gamma$, comparing baseline methods and our proposed approach under an equal number of negative texts. The results highlight the limitations of existing methods, whereas our consistent inter-modal based method effectively mitigates these issues. More results can be found in Appendix \ref{sp::misclassification}. 

\subsection{InterNeg: Inter-modal Guided OOD Detection}

To ensure the distance consistency in OOD detection with CLIP-like VLMs, we propose InterNeg, a simple yet effective method that utilizes consistent inter-modal distance from textual and visual perspectives without requiring training on ID or extra data. 

\subsubsection{Inter-modal Guided Negative Text Selection}
\label{sec:1}
From the textual perspective, we introduce an inter-modal guided strategy for negative text selection. To compute the inter-modal distance, we first need to obtain ID image proxies. Intuitively, we randomly sample $N$ ID images per class from the training set and encode them using the CLIP image encoder $\mathcal{I}(\cdot)$ to obtain their embeddings. Subsequently, we compute the class-wise mean embeddings as ID image proxies $\boldsymbol{p}_i$:
\begin{equation}
    \boldsymbol{p}_i=\frac{1}{N}\sum_{j=1}^{N}\mathcal{I}(\boldsymbol{x}_{ij}), \quad  i=1,2,\cdots,C,
\end{equation}
where $\boldsymbol{x}_{ij}$ denotes the $j$-th sample in the $i$-th ID class. We then define the \textbf{ID inter-modal base distance} for each class with ID text proxies $\boldsymbol{e}_i$:
\begin{equation}
    d_i^{\text{base}} = 1-\text{cos}(\boldsymbol{e}_i,\boldsymbol{p}_i),\quad  i=1,2,\cdots,C,
\end{equation}

Next, we consider using $d_i^{\text{base}}$ to select negative texts. Given a text $y$ from a large-scale corpus database WordNet \cite{fellbaum1998wordnet}, we first compute the inter-modal distance with ID image proxies:
\begin{equation}
\begin{aligned}
    &\boldsymbol{e}^y=\mathcal{T}(\mathcal{E}(\text{prompt}(y))), \\ &d_i(\boldsymbol{e}^y) = 1-cos(\boldsymbol{e}^y,\boldsymbol{p}_i),\quad  i=1,2,\cdots,C,
\end{aligned}
\end{equation}
where $d_i(\boldsymbol{e}^y)$ denotes the inter-modal distance for text $y$ of the $i$-th class and $\mathcal{E}(\cdot)$ is the word embedding function. 

Intuitively, to select texts that effectively discriminate against ID labels, we choose those that satisfy $\forall i,\ d_i(\boldsymbol{e}^y) > d_i^{\text{base}}$ as candidate texts. We refer to such texts as \textbf{inter-modal guided negative texts}. This criterion guarantees that inter-modal guided negative texts maintain large distances from all class-wise ID image-text pairs, improving their ability to approximate true OOD labels. Lastly, the negative set $\mathcal{Y}^-$ is formed by selecting the top-$M$ candidates with the highest \textbf{deviation degree}, which quantifies the discriminative capability of each negative text:
\begin{equation}
\label{eq:degree}
   D(\boldsymbol{e}^y) = \sum_{i=1}^Cd_i(\boldsymbol{e}^y)-d_i^{\text{base}},
\end{equation}

Consequently, the negative texts selected above ensure consistency with the inter-modal optimization goal of CLIP-style VLMs are optimized for, further enhancing OOD detection performance.

\begin{algorithm}[t]
\caption{InterNeg}
\label{al:method}
\begin{algorithmic}[1]
    \Require ID label space $\mathcal{Y}^{\text{ID}}$ and test dataset $\mathcal{X}$;
    \State Select top-$M$ inter-modal guided negative texts $\mathcal{Y}^-$ based on $\mathcal{Y}^{\text{ID}}$ in Section \ref{sec:1};
    \State Initialize an empty extra negative text embeddings set $\mathcal{N}^-$;
    \For {$x \in \mathcal{X}$}
    \State Calculate the OOD score $S(\boldsymbol{x})$ using Eq.(\ref{eq:ood}) ;
    \If {$S(\boldsymbol{x})\leq \beta$}\\
    \textcolor{blue!80!cyan}{$\triangleright$ Step 1: Generating Extra Negative Text Embeddings}  
    \State Invert high-confidence OOD images to generate extra negative text embedding $\boldsymbol{e}_v^-$;\\
    \textcolor{blue!80!cyan}{$\triangleright $ Step 2: Filtering Extra Negative Text Embeddings}  
    \If{$\forall i,\ d_i(\boldsymbol{e}_v^-) > d_i^{\text{base}}$}
    \State Add $\boldsymbol{e}_v^-$ to the set $\mathcal{N}^-$ and record the deviation degree using Eq.(\ref{eq:degree});
    \If{$|\mathcal{N}^-|> K$ }
    \State Only retain the top-$K$ text embeddings in set $\mathcal{N}^-$ based on the deviation degrees; 
    \EndIf
    \State Compute the final OOD score $S_{final}$ again using Eq.(\ref{eq:ood}) ;
    \EndIf
    \EndIf
    \EndFor
    \State \Return Collect all the OOD scores $S_{final}$.
\end{algorithmic} 
\end{algorithm}

\definecolor{top1}{RGB}{212,228,251}
\begin{table*}[t]
\small
\centering
\caption{OOD detection results with ID dataset of ImageNet-1k and traditional Four-OOD datasets using CLIP ViT-B/16 architecture. $\uparrow$ indicates larger values are better and $\downarrow$ indicates smaller values are better. All values are percentages with \textbf{bold} and \underline{underline} indicating the best and second-best results, respectively.}
\label{tab:ood_results}
\begin{adjustbox}{width=\textwidth}
\begin{tabular}{lcccccccccc}
\toprule
\multicolumn{1}{l|}{\multirow{3}{*}{Methods}} & \multicolumn{8}{c|}{OOD Datasets}                                                                                                                                          & \multicolumn{2}{c}{\multirow{2}{*}{Average}} \\
\multicolumn{1}{l|}{}                         & \multicolumn{2}{c}{iNaturalist}     & \multicolumn{2}{c}{SUN}             & \multicolumn{2}{c}{Places}          & \multicolumn{2}{c|}{Textures}                            & \multicolumn{2}{c}{}                         \\ \cmidrule{2-11} 
\multicolumn{1}{l|}{}                         & AUROC$\uparrow$ & FPR95$\downarrow$ & AUROC$\uparrow$ & FPR95$\downarrow$ & AUROC$\uparrow$ & FPR95$\downarrow$ & AUROC$\uparrow$ & \multicolumn{1}{c|}{FPR95$\downarrow$} & AUROC$\uparrow$      & FPR95$\downarrow$     \\ \midrule
\multicolumn{11}{c}{\textbf{Visual-based Methods (requiring training on ID or extra data)}}    \\
\multicolumn{1}{l|}{MSP \cite{DBLP:conf/iclr/HendrycksG17}}                      & 87.44           & 58.36             & 79.73           & 73.72             & 79.67           & 74.41             & 79.69           & \multicolumn{1}{c|}{71.93}            & 81.63                & 69.61                 \\
\multicolumn{1}{l|}{ODIN \cite{DBLP:conf/iclr/LiangLS18}}                    & 94.65           & 30.22             & 87.17           & 54.04             & 85.54           & 55.06             & 87.85           & \multicolumn{1}{c|}{51.67}             & 88.80                & 47.75                 \\
\multicolumn{1}{l|}{Energy \cite{DBLP:conf/nips/LiuWOL20}}                  & 95.33           & 26.12             & 92.66           & 35.97             & 91.41           & 39.87             & 86.76           & \multicolumn{1}{c|}{57.61}             & 91.54                & 39.89                 \\
\multicolumn{1}{l|}{GradNorm \cite{DBLP:conf/nips/HuangGL21}}                 & 72.56           & 81.50             & 72.86           & 82.00             & 73.70           & 80.41             & 70.26           & \multicolumn{1}{c|}{79.36}             & 72.35                & 80.82                 \\
\multicolumn{1}{l|}{ViM \cite{DBLP:conf/cvpr/Wang0F022}}                      & 93.16           & 32.19             & 87.19           & 54.01             & 83.75           & 60.67             & 87.18           & \multicolumn{1}{c|}{53.94}             & 87.82                & 50.20                 \\
\multicolumn{1}{l|}{KNN \cite{DBLP:conf/icml/SunM0L22}}                      & 94.52           & 29.17             & 92.67           & 35.62             & 91.02           & 39.61             & 85.67           & \multicolumn{1}{c|}{64.35}             & 90.97                & 42.19                 \\
\multicolumn{1}{l|}{VOS \cite{DBLP:conf/iclr/DuWCL22}}                      & 94.62           & 28.99             & 92.57           & 36.88             & 91.23           & 38.39             & 86.33           & \multicolumn{1}{c|}{61.02}             & 91.19                & 41.32                 \\ \midrule
\multicolumn{11}{c}{\textbf{VLM-based Methods (requiring training on ID or extra data)}}    \\

\multicolumn{1}{l|}{LoCoOp \cite{DBLP:conf/nips/MiyaiYIA23}}                  & 96.86           & 16.05             & 95.07           & 23.44             & 91.98           & 32.87             & 90.19           & \multicolumn{1}{c|}{42.28}             & 93.52                & 28.66                 \\
\multicolumn{1}{l|}{LSN \cite{DBLP:conf/iclr/NieZ0L0024}}                      & 95.83           & 21.56             & 94.35           & 26.32             & 91.25           & 34.48             & 90.42           & \multicolumn{1}{c|}{38.54}           & 92.96                & 30.22                 \\
\multicolumn{1}{l|}{ID-Like \cite{DBLP:conf/cvpr/BaiHCJHZ24}}                & 98.19           & 8.98             & 91.64           & 42.03             & 90.57           & 44.00            & 94.32           & \multicolumn{1}{c|}{\underline{25.27}}             & 93.68                & 30.07                 \\
\multicolumn{1}{l|}{NegPrompt \cite{DBLP:conf/cvpr/LiP0MZ24}}                & 98.73           & 6.32              & 95.55           & 22.89             & 93.34           & 27.60             & 91.60           & \multicolumn{1}{c|}{35.21}            & 94.81                & 23.01                 \\
\multicolumn{1}{l|}{SUPREME \cite{DBLP:journals/corr/abs-2502-00662}}               & 98.29           & 8.27              & 95.84           & 19.40             & 93.56           & \textbf{26.69}             & 94.45           & \multicolumn{1}{c|}{26.77}             & 95.54                & 20.28                 \\
\multicolumn{1}{l|}{Local-Prompt \cite{localprompt}}               & 98.07           & 8.63             & 95.12           & 23.23             & 92.42           & 31.74             & 92.29           & \multicolumn{1}{c|}{34.50}             & 94.48                & 24.52                 \\
\multicolumn{1}{l|}{ZOC \cite{DBLP:conf/aaai/Esmaeilpour00022}}                      & 86.09           & 87.30             & 81.20           & 81.51             & 83.39           & 73.06             & 76.46           & \multicolumn{1}{c|}{98.90}             & 81.79                & 85.19                 \\
\multicolumn{1}{l|}{CLIPN \cite{DBLP:conf/iccv/WangLYL23}}                    & 95.27           & 23.94             & 93.93           & 26.17             & 92.28           & 33.45             & 90.93           & \multicolumn{1}{c|}{40.83}             & 93.10                & 31.10                 \\
\multicolumn{1}{l|}{LAPT \cite{DBLP:conf/eccv/ZhangZHZ24}}                     & 99.63           & 1.16              & 96.01           & 19.12             & 92.01           & 33.01             & 91.06           & \multicolumn{1}{c|}{40.32}            & 94.68                & 23.40                 \\

\midrule
\multicolumn{11}{c}{\textbf{VLM-based Zero-Shot Methods (no training on ID or extra data)}}    \\

\multicolumn{1}{l|}{Mahalanobis \cite{DBLP:conf/nips/LeeLLS18}}              & 55.89           & 99.33             & 59.94           & 99.41             & 65.96           & 98.54             & 64.23           & \multicolumn{1}{c|}{98.46}             & 61.50                & 98.94                 \\
\multicolumn{1}{l|}{Energy \cite{DBLP:conf/nips/LiuWOL20}}                   & 85.09           & 81.08             & 84.24           & 79.02             & 83.38           & 75.08             & 65.56           & \multicolumn{1}{c|}{93.65}             & 79.57                & 82.21                 \\
\multicolumn{1}{l|}{MCM \cite{DBLP:conf/nips/MingCGSL022}}                      & 94.59           & 32.20             & 92.25           & 38.80             & 90.31           & 46.20             & 86.12           & \multicolumn{1}{c|}{58.50}             & 90.82                & 43.93                 \\
\multicolumn{1}{l|}{EOE \cite{DBLP:conf/icml/CaoZZ0L024}}                     & 97.52           & 12.29             & 95.73           & 20.40             & 92.95           & 30.16             & 85.64           & \multicolumn{1}{c|}{57.53}             & 92.96                & 30.09                 \\
\multicolumn{1}{l|}{NegLabel \cite{DBLP:conf/iclr/Jiang000LZ024}}                 & 99.49           & 1.91              & 95.49           & 20.53             & 91.64           & 35.59             & 90.22           & \multicolumn{1}{c|}{43.56}            & 94.21                & 25.40                 \\

\multicolumn{1}{l|}{CLIPScope \cite{Fu_2025_WACV}}                     & 99.61           & 1.29             & 96.77           & 15.56             & 93.54           & 28.45             & 91.41           &\multicolumn{1}{c|}{38.37}            & 95.30                & 20.88                 \\
\multicolumn{1}{l|}{CoVer \cite{DBLP:conf/nips/ZhangZWL0024}}                     & 95.98           & 22.55             & 93.42           & 32.85            & 90.27           & 40.71             & 90.14           & \multicolumn{1}{c|}{43.39}            & 92.45                & 34.88                 \\
\multicolumn{1}{l|}{CSP \cite{DBLP:conf/nips/Chen0X24}}                     & 99.60           & 1.54             & 96.66           & 13.66            & 92.90           & 29.32             & 93.86           & \multicolumn{1}{c|}{25.52}            & 95.76                & \underline{17.51}                 \\
\multicolumn{1}{l|}{AdaNeg \cite{DBLP:conf/nips/ZhangZ24}}                   & \underline{99.71}           & \underline{0.59}              & \underline{97.44}           & \underline{9.50}              & \underline{94.55}           & 34.34             & \underline{94.93}           & \multicolumn{1}{c|}{31.27}             & \underline{96.66}                & 18.92                 \\
\multicolumn{1}{l|}{\cellcolor{top1}\textbf{InterNeg}}         & \cellcolor{top1}\textbf{99.79}           & \cellcolor{top1}\textbf{0.40}              & \cellcolor{top1}\textbf{98.68}           & \cellcolor{top1}\textbf{6.78}             & \cellcolor{top1}\textbf{95.01}          & \cellcolor{top1}\underline{27.11}          & \cellcolor{top1}\textbf{96.26}           & \multicolumn{1}{c|}{\cellcolor{top1}\textbf{21.85}}             & \cellcolor{top1}\textbf{97.43}                & \cellcolor{top1}\textbf{14.04}                 \\ \bottomrule
\end{tabular}
\end{adjustbox}
\end{table*}

\subsubsection{Inter-modal Guided Extra Negative Texts}
\label{sec:2}

From the visual perspective, we dynamically leverage OOD images to enhance the negative textual space through inter-modal distance during inference. To obtain these OOD images, we use test samples that are classified as OOD with high confidence. These high-confidence OOD images are then inverted into the textual space to produce extra negative text embeddings. Formally, let $\mathcal{N}^-$ represent the set of extra negative text embeddings, where $|\mathcal{N}^-|$ denotes the number of embeddings in the set. The enhanced OOD score function is then defined as follows:

\definecolor{ID}{RGB}{204,96,96}    
\definecolor{SNT}{RGB}{100,180,130}
\definecolor{ENT}{RGB}{80,170,210}  
\begin{equation}
\label{eq:ood}
        \begin{aligned}
    S(\boldsymbol{x})=&\left({\color{ID}\sum_{i=1}^C e^{\text{cos}(\boldsymbol{h},\boldsymbol{e}_i)/\tau}}\right) \left({\color{ID}\underbrace{\sum\nolimits_{i=1}^C e^{\text{cos}(\boldsymbol{h},\boldsymbol{e}_i)/\tau}}_{\text{ID Labels}}}+ \right.\\
    &\left.{\color{SNT}\underbrace{\sum\nolimits_{i=1}^M e^{\text{cos}(\boldsymbol{h},\boldsymbol{e}^-_i)/\tau}}_{\text{Selected Negative Texts}}}+{\color{ENT}\underbrace{\sum\nolimits_{\boldsymbol{e}_v^-\in \mathcal{N}^-}^{|\mathcal{N}^-|} e^{\text{cos}(\boldsymbol{h},\boldsymbol{e}^-_v)/\tau}}_{\text{Extra Negative Text Embeddings}}}\right)^{-1},
        \end{aligned}
\end{equation}
where $\boldsymbol{e}_v^-$ denotes the extra negative text embedding, whose details are described below. Initially, $\mathcal{N}^-$ is set to be empty. During inference, our method dynamically expands $\mathcal{N}^-$ through an iterative two-step process.

\paragraph{Step 1: Generating Extra Negative Text Embeddings.}
If $S(\boldsymbol{x}) \leq \beta$, the image $\boldsymbol{x}$ is identified as a high-confidence OOD image, where $\beta$ is a hyperparameter serving as the threshold for this determination. To obtain the negative text embeddings, we use the modality inversion technique in \cite{crossthegap,DBLP:conf/iccv/BaldratiA0B23} to transform the above high-confidence OOD images
into textual space. Specifically, for a high-confidence OOD image $\boldsymbol{x}$, we first randomly initialize a set of $T$ pseudo-tokens $\boldsymbol{v}=\{v_1,v_2,\cdots,v_T\}$ and concatenate with the text template (\textit{e.g.}, "a photo of") to form $\boldsymbol{\bar{v}}=[\mathcal{E}(''a\; photo\; of''),\boldsymbol{v}]$. Then, we extract the image embedding $\boldsymbol{h}$ and text embedding $\boldsymbol{e}_v^-$ and as follows:
\begin{equation}
    \boldsymbol{e}_v^-=\mathcal{T}(\boldsymbol{\bar{v}}),\quad \boldsymbol{h}=\mathcal{I}(\boldsymbol{x})
\end{equation}
Next, we optimize $\boldsymbol{v}$ through minimizing the cosine distance loss between the image and text embeddings $\mathcal{L}=1-cos(\boldsymbol{e}_v^-,\boldsymbol{h})$, which is based on the inter-modal distance. Hence, \textbf{we get an extra negative text embedding} $\boldsymbol{e}_v^-$. Please refer to Algorithm \ref{al:inversion} for the pseudo-code of the modality inversion process.

\paragraph{Step 2: Filtering Extra Negative Text Embeddings.}
Since a fixed threshold is applied above, some noisy images may be incorrectly included. To mitigate these noisy images, we implement an \textbf{inter-modal guided dynamic filtering mechanism}.
On one hand, we also leverage an \textbf{inter-modal guided selection} for the generated negative text embeddings. Specifically, we utilize the extra negative text embedding $\boldsymbol{e}_v^-$ to calculate $d_i(\boldsymbol{e}_v^-)$ and compare with $d_i^{\text{base}}$ (as described in Section \ref{sec:1}). Only those $\boldsymbol{e}_v^-$ that satisfy $\forall i,\ d_i(\boldsymbol{e}_v^-) > d_i^{\text{base}}$ are added to $\mathcal{N}^-$ with their corresponding deviation degrees $ D(\boldsymbol{e}_v^-)$, others are discarded. On the other hand, we \textbf{limit the size of the extra negative text embeddings set} $\mathcal{N}^-$ to a maximum of $K$, where $K$ is a hyperparameter. Specifically, when the addition of new embeddings causes $|\mathcal{N}^-|$ to exceed $K$, we sort all embeddings in $\mathcal{N}^-$ by their deviation degrees and retain only the top-$K$ highest-scoring ones, discarding the rest. The overall pipeline of our method is summarized in Algorithm \ref{al:method}.


\section{Experiments}

\begin{table}[t]
\footnotesize
\centering
\caption{OOD detection results with ID dataset of ImageNet-1k and OpenOOD benchmark using CLIP ViT-B/16 architecture. Full results are available in Table \ref{tab:full_imagenet}.}
\label{tab:ood_results_openood}
\begin{tabular}{lcccc}
\toprule
\multicolumn{1}{l|}{\multirow{2}{*}{Methods}} & \multicolumn{2}{c|}{FPR95 $\downarrow$} & \multicolumn{2}{c}{AUROC $\uparrow$}  \\ \cmidrule{2-5}  
\multicolumn{1}{l|}{} & Near-OOD  & \multicolumn{1}{c|}{Far-OOD}  & Near-OOD & Far-OOD  \\
\midrule
\multicolumn{5}{c}{\textcolor{gray}{\textbf{Methods requiring training on ID or extra data}}} \\

\multicolumn{1}{l|}{\textcolor{gray}{GEN \cite{DBLP:conf/cvpr/LiuLZ23}}} & \textcolor{gray}{–} & \multicolumn{1}{c|}{\textcolor{gray}{–}} & \textcolor{gray}{78.97} & \textcolor{gray}{90.98}  \\
\multicolumn{1}{l|}{\textcolor{gray}{ReAct \cite{DBLP:conf/nips/SunGL21}}} & \textcolor{gray}{–} & \multicolumn{1}{c|}{\textcolor{gray}{–}} & \textcolor{gray}{79.94} & \textcolor{gray}{93.70}  \\
\multicolumn{1}{l|}{\textcolor{gray}{RMDS \cite{DBLP:journals/corr/abs-2106-09022}}}  & \textcolor{gray}{–} & \multicolumn{1}{c|}{\textcolor{gray}{–}} & \textcolor{gray}{80.09} & \textcolor{gray}{92.60} \\
\multicolumn{1}{l|}{\textcolor{gray}{SCALE \cite{DBLP:conf/iclr/XuCFY24}}} & \textcolor{gray}{–} & \multicolumn{1}{c|}{\textcolor{gray}{–}} & \textcolor{gray}{81.36} & \textcolor{gray}{96.53}  \\
\multicolumn{1}{l|}{\textcolor{gray}{ASH \cite{DBLP:conf/iclr/DjurisicBAL23}}} & \textcolor{gray}{\textcolor{gray}{–}} & \multicolumn{1}{c|}{\textcolor{gray}{–}} & \textcolor{gray}{82.16} & \textcolor{gray}{96.05}  \\
\multicolumn{1}{l|}{\textcolor{gray}{LAPT \cite{DBLP:conf/eccv/ZhangZHZ24}}} & \textcolor{gray}{58.94} & \multicolumn{1}{c|}{\textcolor{gray}{24.86}} & \textcolor{gray}{82.63} & \textcolor{gray}{94.26}   \\
\midrule
\multicolumn{5}{c}{\textbf{Zero-shot methods (no training on ID or extra data)}} \\

\multicolumn{1}{l|}{MCM \cite{DBLP:conf/nips/MingCGSL022}} & 79.02 & \multicolumn{1}{c|}{68.54} & 60.11 & 84.77\\
\multicolumn{1}{l|}{NegLabel \cite{DBLP:conf/iclr/Jiang000LZ024}} & 69.45 & \multicolumn{1}{c|}{23.73} & 75.18 & 94.85  \\
\multicolumn{1}{l|}{AdaNeg \cite{DBLP:conf/nips/ZhangZ24}}  & 67.51 & \multicolumn{1}{c|}{17.31} & 76.70 & 96.43 \\
\multicolumn{1}{l|}{\cellcolor{top1}\textbf{InterNeg}}  & \cellcolor{top1}\textbf{65.43} & \multicolumn{1}{c|}{\cellcolor{top1}\textbf{16.96}} & \cellcolor{top1}\textbf{82.20} & \cellcolor{top1}\textbf{96.71} \\
\bottomrule
\end{tabular}
\end{table}

\subsection{Main Results}
\paragraph{OOD Detection Performance Comparison on Traditional Four-OOD Datasets.} In Table \ref{tab:ood_results}, we compare our proposed method with other existing OOD detection methods. Specifically, we report the traditional visual-based methods from MSP \cite{DBLP:conf/iclr/HendrycksG17} to VOS, and recent VLM-based methods from LoCoOp \cite{DBLP:conf/nips/MiyaiYIA23} to AdaNeg \cite{DBLP:conf/nips/ZhangZ24}. The experimental results demonstrate that our proposed method achieves state-of-the-art performance, significantly outperforming the closest baseline with substantial improvements of 0.77\% in AUROC and 3.47\% in FPR95, thereby validating the effectiveness and superiority of our method.

\paragraph{OOD Detection Performance Comparison on OpenOOD Setup.} Following \cite{DBLP:conf/nips/ZhangZ24}, we also evaluate our method on the OpenOOD benchmark, which contains challenging Near-OOD and easy Far-OOD datasets. Referred from \cite{DBLP:conf/nips/ZhangZ24,DBLP:conf/nips/YangWZZDPWCLSDZ22,DBLP:journals/corr/abs-2306-09301}, we report the results of zero-shot methods and methods requiring training on ID or extra data in Table \ref{tab:ood_results_openood}. Based on the experimental results, our proposed method demonstrates consistent superiority over existing zero-shot approaches across both Near-OOD and Far-OOD scenarios. Notably, in Near-OOD scenarios, our approach achieves significant improvements with a 5.50\% increase in AUROC and a 2.09\% reduction in FPR95. These substantial enhancements make our method competitive with approaches that require training on any data.

\definecolor{Intra}{RGB}{156,148,235}
\definecolor{Inter}{RGB}{58,194,210} 
\begin{table}[t]
  \centering
  \caption{Ablation study. Average results on the Four-OOD benchmark, where SNT denotes \underline{S}elected \underline{N}egative \underline{T}exts and ENT denotes \underline{E}xtra \underline{N}egative \underline{T}ext embeddings, both of which can be obtained by \textcolor{Intra}{Intra}-modal or \textcolor{Inter}{Inter}-modal distances.}
  \label{tab:ablation}
  \begin{adjustbox}{width=0.98\linewidth}
  \begin{tabular}{cccc|cc}
  \toprule
  \color{Intra}Intra SNT    &  \color{Inter}Inter SNT & \color{Intra}Intra ENT&  \color{Inter}Inter ENT     & AUROC $\uparrow$ & FPR95 $\downarrow$ \\ \midrule
  \checkmark &      &    &    & 94.21  & 25.40                  \\
   & \checkmark &       &    & 94.56  & 24.12  \\ \midrule
  \checkmark&    & \checkmark     &    & 95.89 & 	20.84    \\
  &   \checkmark &  \checkmark &       & 96.22 &19.99     \\
  \checkmark&    &   & \checkmark      & 97.11 &14.76     \\
  & \checkmark & &\checkmark & 97.43 &14.04                    \\ \bottomrule
  \end{tabular}
  \end{adjustbox}
\end{table}

\begin{figure}[t]
  \centering
  \begin{subfigure}{0.23\textwidth}
    \centering
    \includegraphics[width=\linewidth]{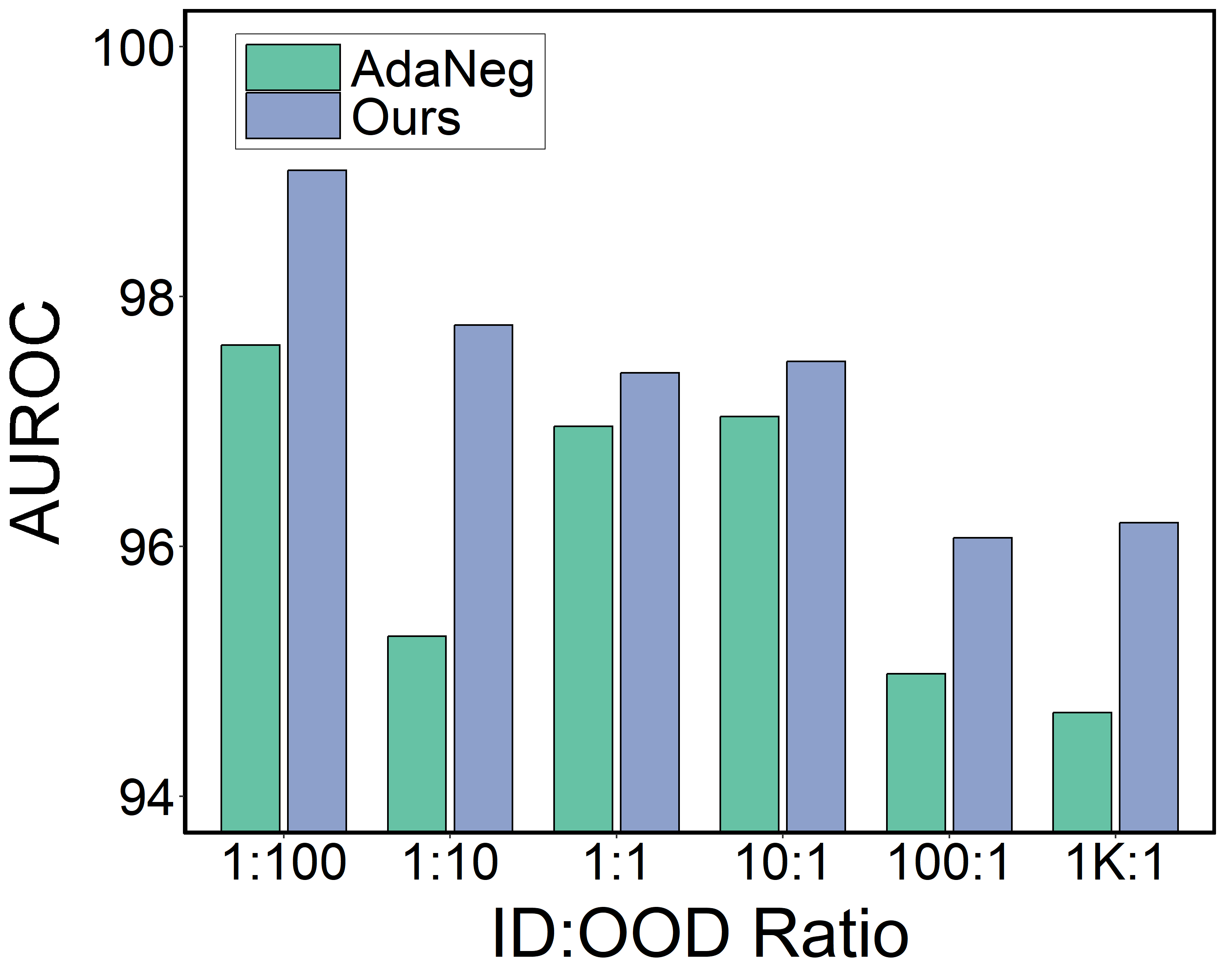}
    \caption{AUROC $\uparrow$}
    \label{fig:imbalance_auc}
  \end{subfigure}
\hspace{0.01cm}
  \begin{subfigure}{0.23\textwidth}
    \centering
    \includegraphics[width=\linewidth]{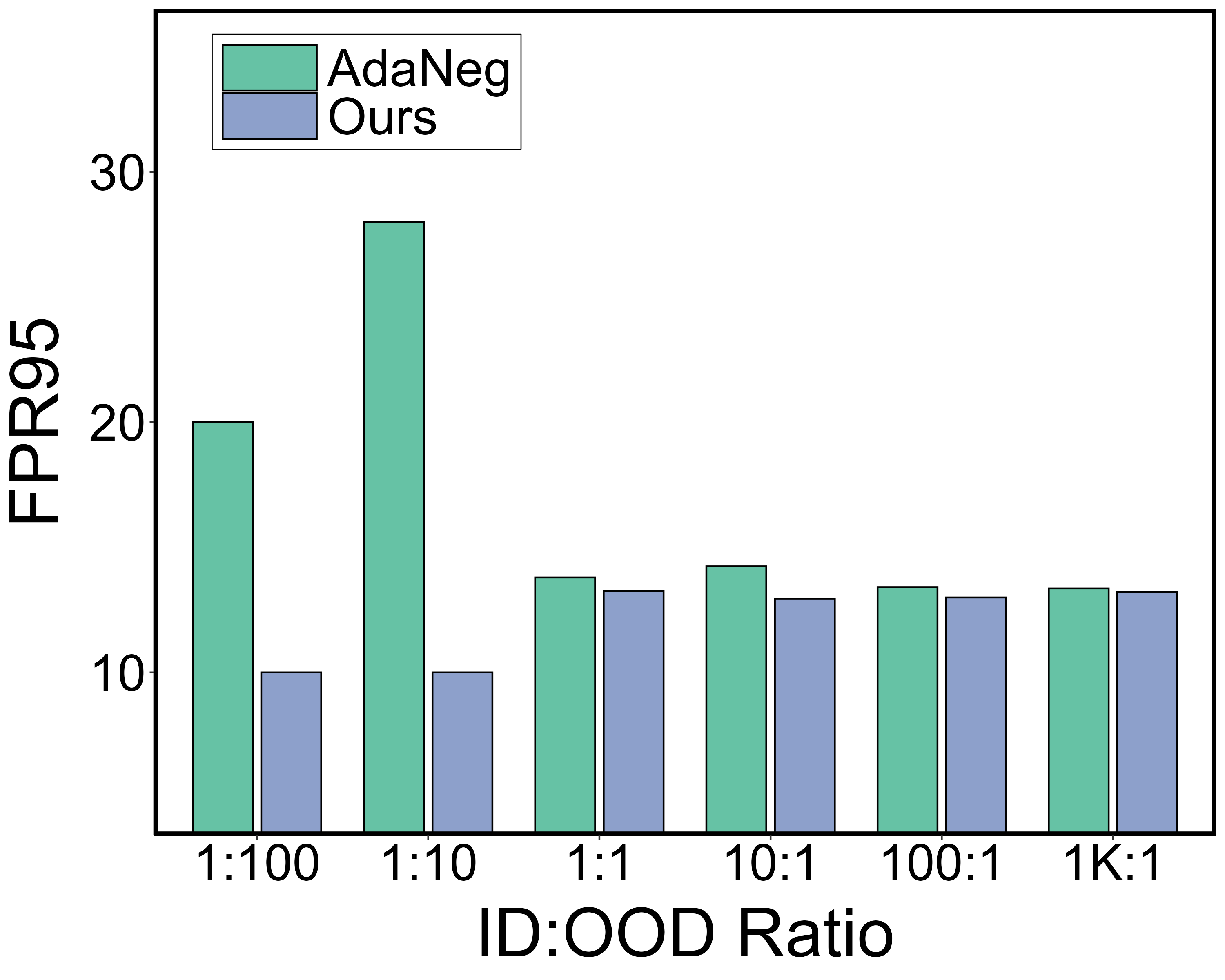} 
    \caption{FPR95 $\downarrow$}
    \label{fig:imbalance_fpr}
  \end{subfigure}
  \caption{AUROC $\uparrow$ and FPR95 $\downarrow$ average performance under varying ID:OOD ratios.}
  \label{fig:imbalance}
\end{figure}

\subsection{Ablation Study and Discussion}

\paragraph{Ablation on Each Module in InterNeg.} Table \ref{tab:ablation} presents a comprehensive ablation study comparing our proposed method with four alternative approaches to validate its effectiveness: (1) \textcolor{Intra}{Intra SNT}, this baseline utilizes negative texts selected by intra-modal distance in NegLabel \cite{DBLP:conf/iclr/Jiang000LZ024}. (2) \textcolor{Inter}{Inter SNT}, it represents the first component of our proposed method (Section \ref{sec:1}), employing inter-modal guided negative texts. (3) \textcolor{Intra}{Intra SNT} + \textcolor{Intra}{Intra ENT}, a variant where both components of our method are replaced with intra-modal distance. The selected negative texts are intra-modal guided, and extra negative text embeddings are filtered using intra-modal distance. (4) \textcolor{Inter}{Inter SNT} + \textcolor{Intra}{Intra ENT}, a hybrid approach that uses our inter-modal guided negative texts selection but employs intra-modal distance for filtering extra negative text embeddings. (5) \textcolor{Intra}{Intra SNT} + \textcolor{Inter}{Inter ENT}, a hybrid approach combining NegLabel with our proposed inter-modal guided extra negative text embeddings. The experimental results demonstrate the superior performance of both components in our method (\textcolor{Inter}{Inter SNT} + \textcolor{Inter}{Inter ENT}).

\paragraph{Imbalanced ID and OOD Test Data.} To evaluate our method under imbalanced ID and OOD test settings \cite{DBLP:journals/pami/WangXYXZCH26,NEURIPS2023_973a0f50,DirMixE,focalsam,10.5555/3692070.3694407,pmlr-v235-zhao24o,aucseg}, we construct datasets with varying ID:OOD ratios. Specifically, we first create four basic configurations by: (1) randomly selecting 1K samples from the SUN dataset as OOD samples, and (2) independently sampling 10, 100, 1K, and 10K images from the ImageNet-1K dataset as ID samples. This procedure yields ID:OOD ratios of 1:100, 1:10, 1:1, and 10:1 respectively. For more extreme ratios, we generate additional configurations by: (1) selecting only 10 samples from the SUN dataset as OOD data, and (2) pairing them with 1K or 1K ImageNet-1K samples to achieve ID:OOD ratios of 100:1 and 1K:1. All experimental data are drawn from Four-OOD benchmark to ensure no overlap between ID and OOD samples.

As shown in Figure \ref{fig:imbalance}, we compare our method with the best baseline AdaNeg (with adaptive gap strategy). Notably, even under highly imbalanced ID/OOD data distributions, our approach consistently outperforms AdaNeg by a substantial margin. This performance gap validates that our inter-modal guided dynamic filtering mechanism effectively mitigates the noise introduced by the fixed threshold, thereby confirming the robustness and reliability of our proposed method.

Further analyses regarding ID/OOD dataset ordering, various CLIP architectures, cross-domain settings, inference cost, and sensitivity to corpus choice are provided in Appendix \ref{sp:additional}.

\begin{figure}[t]
    \centering
    \begin{subfigure}[b]{0.49\linewidth}
        \includegraphics[width=\linewidth]{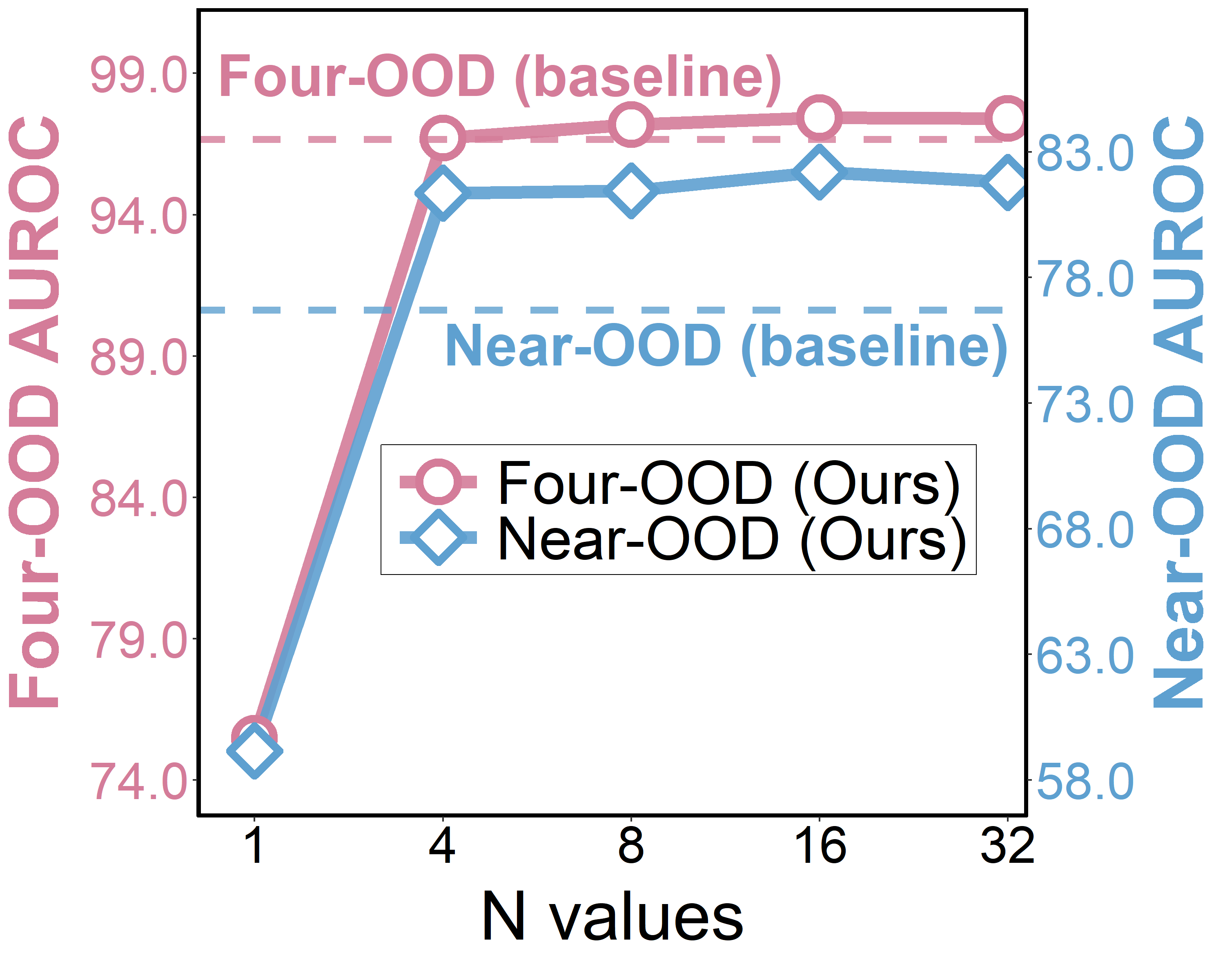}
        \caption{ID images per class}
        \label{fig:analysis-n}
    \end{subfigure}
    \hfill
    \begin{subfigure}[b]{0.49\linewidth}
        \includegraphics[width=\linewidth]{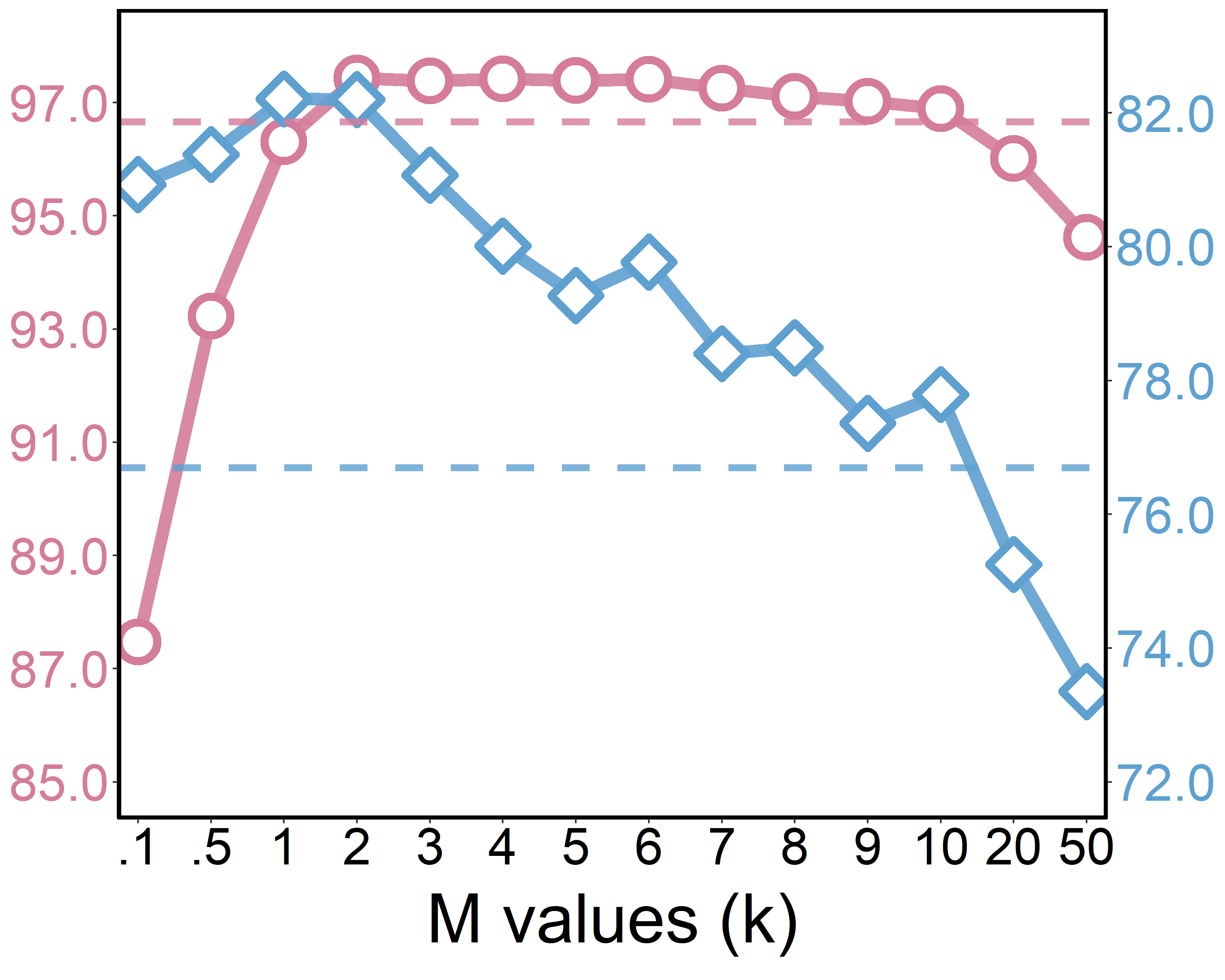}
        \caption{Selected negative texts}
        \label{fig:analysis-m}
    \end{subfigure}

    \vspace{0.2cm}

    \begin{subfigure}[b]{0.49\linewidth}
        \includegraphics[width=\linewidth]{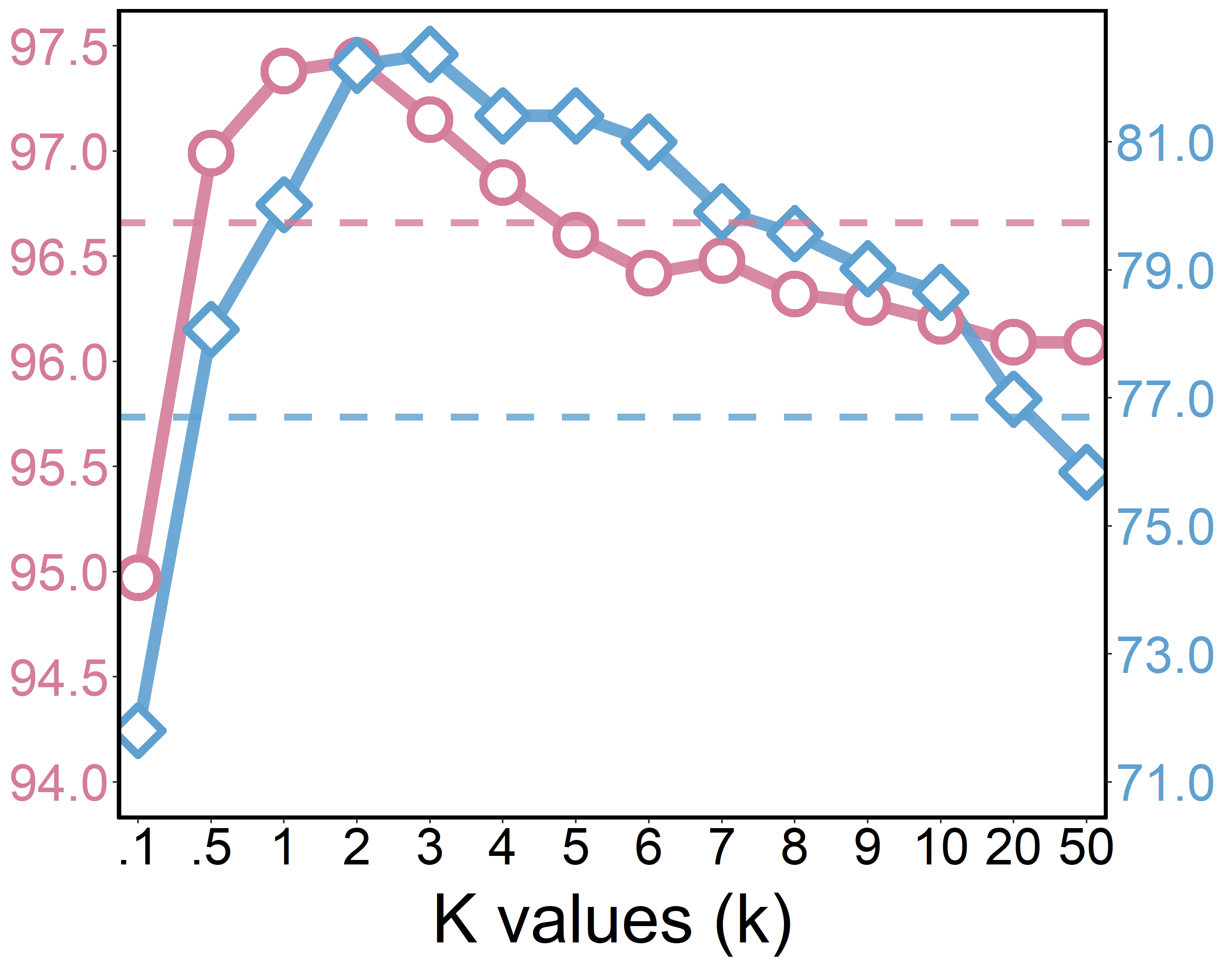}
        \caption{Extra negative text embeddings}
        \label{fig:analysis-k}
    \end{subfigure}
    \hfill
    \begin{subfigure}[b]{0.49\linewidth}
        \includegraphics[width=\linewidth]{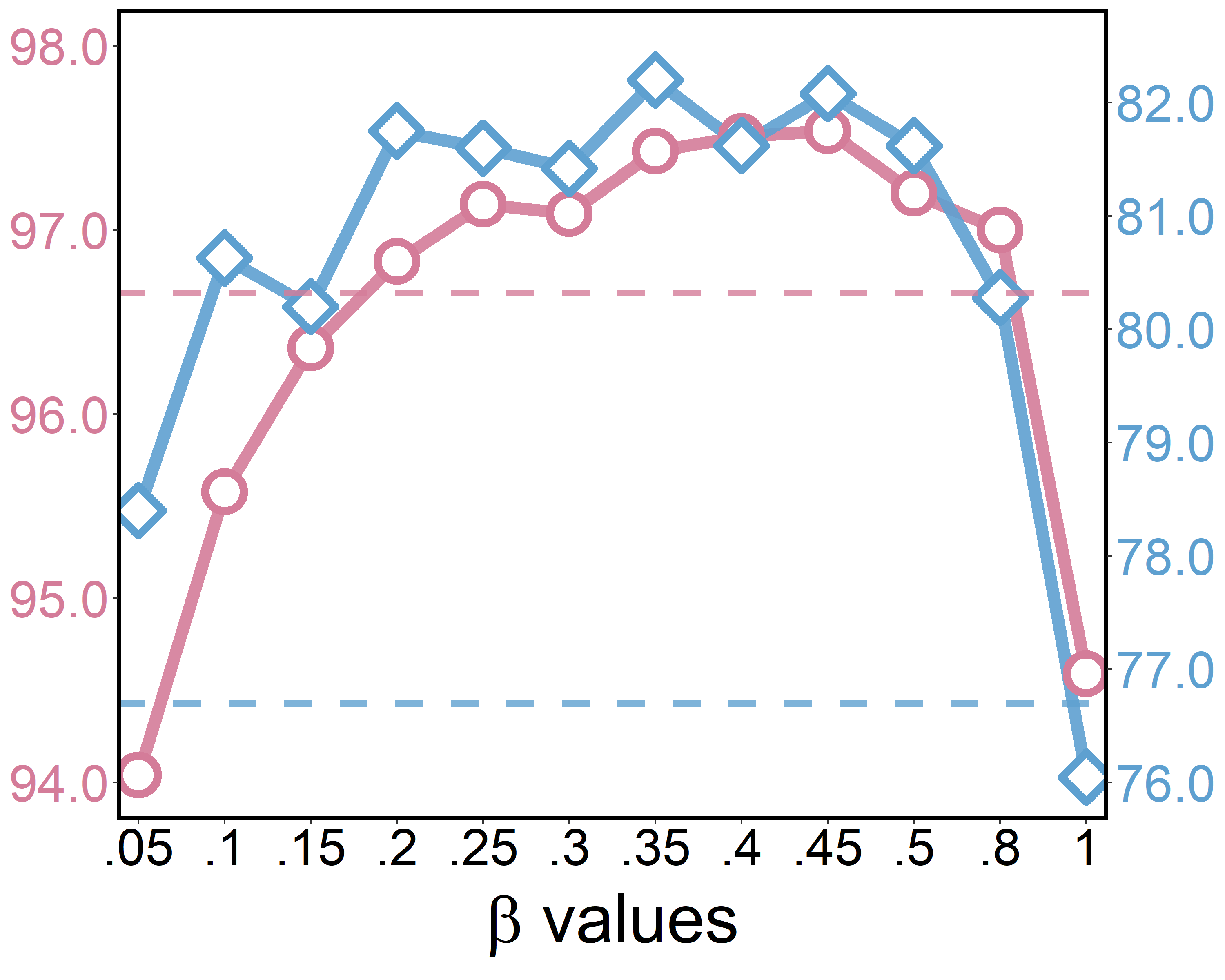}
        \caption{High-confidence OOD threshold}
        \label{fig:analysis-beta}
    \end{subfigure}
    
    \caption{Parameter sensitivity analysis of four key hyperparameters: the number of ID images per class $N$, the number of selected negative texts $M$, the maximum size of extra negative text embeddings set $K$, and the high-confidence OOD threshold $\beta$, evaluated on both Four-OOD and Near-OOD benchmarks using ImageNet-1K as the ID dataset.}
    \label{fig:analysis}
\end{figure}

\subsection{Parameter Sensitivity Analysis}
\paragraph{The Number of ID images Per Class $N$.} To examine the impact of the number of ID images per class serving as image proxies, we present the analysis in Figure \ref{fig:analysis-n}. The results demonstrate that using only one image per class yields suboptimal performance, primarily because a single image may not sufficiently represent the class as an effective proxy. However, when the number of images per class reaches or exceeds 4 ($N\geq4$), our method achieves state-of-the-art performance compared to the closest baseline. This finding indicates that our method requires only a small number of ID images per class to serve as effective image proxies. In all experiments, we choose $N=16$ since it achieves optimal performance.

\paragraph{Selected Negative Texts Number $M$.} As demonstrated in Figure \ref{fig:analysis-m}, our method achieves consistent and superior performance over all baselines across a broad spectrum of $M$ values, highlighting its robustness. 
Empirical results indicate that optimal performance is achieved with a moderate number (\textit{e.g.,} $2000\leq M\leq 9000$) of the selected negative texts. Based on this analysis, we set $M=2000$ in all subsequent experiments.

\paragraph{Maximum Size of Extra Negative Text Embeddings Set $K$.}
As illustrated in Figure \ref{fig:analysis-k}, the model achieves optimal performance when the $K$ value approaches the predefined number of selected negative texts ($M=2000$). Beyond this point, performance gradually degrades as $K$ increases, which can be attributed to the fixed threshold permitting more potentially erroneous OOD samples to be included in the consideration. Thus, we adopt $K=2000$ throughout our experiments as it provides optimal performance for both Four-OOD and Near-OOD benchmarks.

\paragraph{High-confidence OOD Threshold $\beta$.} As shown in Figure \ref{fig:analysis-beta}, our method maintains robust performance across a wide range of $\beta$ values, further validating the effectiveness of the inter-modal guided dynamic filtering mechanism.  Since we choose $K=2000$, even for larger $\beta$ values, our approach selectively retains only the top-$K$ most consistent negative text embeddings, ensuring stable and effective performance. Hence, we choose a moderate value of $\beta=0.35$ for all experiments. 

\section{Conclusion}
\label{sec:conclude}
In this paper, we present InterNeg, a simple yet effective approach that improves OOD detection through consistent inter-modal distance without requiring training on ID or extra data. On one hand, we conduct an inter-modal guided negative text selection from the textual space. On the other hand, we utilize inter-modal guided high-confidence OOD images inversion to generate extra negative text embeddings from the visual space. Extensive experiments across diverse benchmarks demonstrate that InterNeg consistently achieves state-of-the-art performance, confirming our proposed method's effectiveness and robustness.

\section*{Acknowledgement}
This work was supported in part by National Natural Science Foundation of China: 62525212, U23B2051, 62236008, 62441232, 62521007, 62502500, 62576332, and U21B2038, in part by Youth Innovation Promotion Association CAS, in part by the Strategic Priority Research Program of the Chinese Academy of Sciences, Grant No. XDB0680201, in part by the project ZR2025ZD01 supported by Shandong Provincial Natural Science Foundation, in part by the China National Postdoctoral Program for Innovative Talents under Grant BX20240384, in part by Beijing Natural Science Foundation under Grant No. L252144, in part by General Program of the Chinese Postdoctoral Science Foundation under Grant No. 2025M771558, in part by the Beijing Major Science and Technology Project under Contract No. Z251100008125059, and in part by Beijing Academy of Artificial Intelligence (BAAI).
{
    \small
    \bibliographystyle{ieeenat_fullname}
    \bibliography{main}
}

\clearpage
\onecolumn
\setcounter{page}{1}
\appendix
\vspace{2em}
\begin{center}
    \Large \bfseries Appendix Table of Contents
\end{center}
\vspace{1em}

\startcontents[sections]
\printcontents[sections]{l}{1}{\setcounter{tocdepth}{2}}
\vspace{2em}
\clearpage

\section{Pseudo-code for Modality Inversion}
Here, we provide the pseudo-code for the modality inversion process in Algorithm \ref{al:inversion}. This procedure aims to transform a high-confidence OOD image into an extra negative text embedding by optimizing a set of pseudo-tokens. 
\begin{algorithm}[h]
\caption{Modality Inversion From Image To Text \cite{crossthegap}}
\label{al:inversion}
\begin{algorithmic}[1]
    \Require High-confidence OOD image $\boldsymbol{x}$, number of pseudo-tokens $T$, number of optimization steps $S$;
    \State Initialize $\boldsymbol{v}=\{v_1,v_2,\cdots,v_T\}$;
    \State Extract image embedding $\boldsymbol{h}=\mathcal{I}(\boldsymbol{x})$;
    \For {$s=1$ to $S$}
    \State Form $\boldsymbol{\bar{v}}=[\mathcal{E}(''a\; photo\; of''),\boldsymbol{v}]$;
    \State Extract text embedding $ \boldsymbol{e}_v^-=\mathcal{T}(\boldsymbol{\bar{v}})$;
    \State Calculate the cosine loss $\mathcal{L}=1-cos(\boldsymbol{e}_v^-,\boldsymbol{h})$;
    \State Update $\boldsymbol{v}$ to minimize $\mathcal{L}$;
    \EndFor
    \State \Return Negative text embedding $ \boldsymbol{e}_v^-=\mathcal{T}(\boldsymbol{\bar{v}})$.
\end{algorithmic} 
\end{algorithm} 

\section{Additional Results}
\label{sp:additional}
\subsection{Full results on the OpenOOD Benchmark}
\begin{table}[h]
\small
\centering
\caption{Full results of our method with ID dataset of ImageNet-1K on the OpenOOD benchmark.}
\label{tab:full_imagenet}
\begin{tabular}{l|l|c|c}
\toprule
Near / Far OOD & Datasets & FPR95 $\downarrow$ & AUROC $\uparrow$ \\
\midrule
\multirow{3}{*}{Near-OOD} & SSB-hard \cite{DBLP:conf/iclr/Vaze0VZ22} & 69.96 & 80.24 \\
 & NINCO \cite{DBLP:conf/icml/BitterwolfM023} & 60.90  & 84.16 \\
 & \textbf{Mean} & \textbf{65.43} & \textbf{82.20} \\
\midrule
\multirow{4}{*}{Far-OOD} & iNaturalist \cite{DBLP:conf/cvpr/HornASCSSAPB18} & 0.40 & 99.71 \\
 & Textures \cite{DBLP:conf/cvpr/CimpoiMKMV14} & 18.17 & 96.74 \\
 & OpenImage-O \cite{DBLP:conf/cvpr/Wang0F022} & 32.30 & 93.69 \\
 & \textbf{Mean} & \textbf{16.96} & \textbf{96.71} \\
\bottomrule
\end{tabular}
\end{table}

\subsection{More Results on the OpenOOD Benchmark}
\label{sp:openood}

\begin{table}[H]
\small
\centering
\caption{OOD detection results with ID dataset of CIFAR100 on the OpenOOD benchmark using CLIP ViT-B/16 architecture. Full results are available in Table \ref{tab:full_cifar100}.}
\label{tab:ood_results_openood_cifar100}
\begin{tabular}{lcccc}
\toprule
\multicolumn{1}{l|}{\multirow{2}{*}{Methods}} & \multicolumn{2}{c|}{FPR95 $\downarrow$} & \multicolumn{2}{c}{AUROC $\uparrow$}  \\ \cmidrule{2-5}  
\multicolumn{1}{l|}{} & Near-OOD  & \multicolumn{1}{c|}{Far-OOD}  & Near-OOD & Far-OOD  \\
\midrule
\multicolumn{5}{c}{\textcolor{gray}{\textbf{Methods requiring training on ID or extra data}}} \\

\multicolumn{1}{l|}{\textcolor{gray}{GEN \cite{DBLP:conf/cvpr/LiuLZ23}}} & \textcolor{gray}{–} & \multicolumn{1}{c|}{\textcolor{gray}{–}} & \textcolor{gray}{81.31} & \textcolor{gray}{79.68}  \\
\multicolumn{1}{l|}{\textcolor{gray}{VOS \cite{DBLP:conf/iclr/DuWCL22} + EBO \cite{DBLP:conf/nips/LiuWOL20}}} & \textcolor{gray}{–} & \multicolumn{1}{c|}{\textcolor{gray}{–}} & \textcolor{gray}{80.93} & \textcolor{gray}{81.32}  \\
\multicolumn{1}{l|}{\textcolor{gray}{SCALE \cite{DBLP:conf/iclr/XuCFY24}}}  & \textcolor{gray}{–} & \multicolumn{1}{c|}{\textcolor{gray}{–}} & \textcolor{gray}{80.99} & \textcolor{gray}{81.42} \\
\multicolumn{1}{l|}{\textcolor{gray}{OE \cite{DBLP:conf/iclr/HendrycksMD19} + MSP \cite{DBLP:conf/iclr/HendrycksG17}}}  & \textcolor{gray}{–} & \multicolumn{1}{c|}{\textcolor{gray}{–}} & \textcolor{gray}{88.30} & \textcolor{gray}{81.41} \\
\midrule
\multicolumn{5}{c}{\textbf{Zero-shot methods (no training on ID or extra data)}} \\

\multicolumn{1}{l|}{MCM \cite{DBLP:conf/nips/MingCGSL022}} & 75.20 & \multicolumn{1}{c|}{59.32} & 71.00 & 76.00\\
\multicolumn{1}{l|}{NegLabel \cite{DBLP:conf/iclr/Jiang000LZ024}} & 71.44 & \multicolumn{1}{c|}{40.92} & 70.58 & 89.68  \\
\multicolumn{1}{l|}{AdaNeg \cite{DBLP:conf/nips/ZhangZ24}}  & \textbf{59.07} & \multicolumn{1}{c|}{29.35} & 84.60 & 95.25 \\
\multicolumn{1}{l|}{\cellcolor{top1}\textbf{InterNeg}}  & \cellcolor{top1}62.54 & \multicolumn{1}{c|}{\cellcolor{top1}\textbf{20.02}} & \cellcolor{top1}\textbf{85.45} & \cellcolor{top1}\textbf{96.39} \\
\bottomrule
\end{tabular}
\end{table}

\begin{table}[h!]
\small
\centering
\caption{Full results of our method with ID dataset of CIFAR100 on the OpenOOD benchmark.}
\label{tab:full_cifar100}
\begin{tabular}{l|l|c|c}
\toprule
Near / Far OOD & Datasets & FPR95 $\downarrow$ & AUROC $\uparrow$ \\
\midrule
\multirow{3}{*}{Near-OOD} & CIFAR10 \cite{Krizhevsky2009LearningML} & 60.10 & 84.19 \\
 & TIN \cite{Le2015TinyIV} & 64.99 & 86.71 \\
 & \textbf{Mean} & \textbf{62.54} & \textbf{85.45} \\
\midrule
\multirow{5}{*}{Far-OOD} & MNIST \cite{mnist} &  0.01 & 99.97 \\
 & SVHN \cite{svhn} & 3.23 & 99.41 \\
 & Texture \cite{DBLP:conf/cvpr/CimpoiMKMV14} & 21.16 & 96.84 \\
 & Places365 \cite{DBLP:journals/pami/ZhouLKO018} & 55.68 & 89.36 \\
 & \textbf{Mean} & \textbf{20.02} & \textbf{96.39} \\
\bottomrule
\end{tabular}
\end{table}

\begin{table}[h!]
\small
\centering
\caption{OOD detection results with ID dataset of CIFAR10 on the OpenOOD benchmark using CLIP ViT-B/16 architecture. Full results are available in Table \ref{tab:full_cifar10}.}
\label{tab:ood_results_openood_cifar10}
\begin{tabular}{lcccc}
\toprule
\multicolumn{1}{l|}{\multirow{2}{*}{Methods}} & \multicolumn{2}{c|}{FPR95 $\downarrow$} & \multicolumn{2}{c}{AUROC $\uparrow$}  \\ \cmidrule{2-5}  
\multicolumn{1}{l|}{} & Near-OOD  & \multicolumn{1}{c|}{Far-OOD}  & Near-OOD & Far-OOD  \\
\midrule
\multicolumn{5}{c}{\textcolor{gray}{\textbf{Methods requiring training on ID or extra data}}} \\

\multicolumn{1}{l|}{\textcolor{gray}{PixMix \cite{DBLP:conf/cvpr/HendrycksZMTLSS22} + KNN \cite{DBLP:conf/icml/SunM0L22}}} & \textcolor{gray}{–} & \multicolumn{1}{c|}{\textcolor{gray}{–}} & \textcolor{gray}{93.10} & \textcolor{gray}{95.94}  \\
\multicolumn{1}{l|}{\textcolor{gray}{OE \cite{DBLP:conf/iclr/HendrycksMD19} + MSP \cite{DBLP:conf/iclr/HendrycksG17}}} & \textcolor{gray}{–} & \multicolumn{1}{c|}{\textcolor{gray}{–}} & \textcolor{gray}{94.82} & \textcolor{gray}{96.00}  \\
\multicolumn{1}{l|}{\textcolor{gray}{PixMix \cite{DBLP:conf/cvpr/HendrycksZMTLSS22} + RotPred \cite{DBLP:conf/nips/HendrycksMKS19}}}  & \textcolor{gray}{–} & \multicolumn{1}{c|}{\textcolor{gray}{–}} & \textcolor{gray}{94.86} & \textcolor{gray}{98.18} \\
\midrule
\multicolumn{5}{c}{\textbf{Zero-shot methods (no training on ID or extra data)}} \\

\multicolumn{1}{l|}{MCM \cite{DBLP:conf/nips/MingCGSL022}} & 30.86 & \multicolumn{1}{c|}{17.99} & 91.92 & 95.54\\
\multicolumn{1}{l|}{NegLabel \cite{DBLP:conf/iclr/Jiang000LZ024}} & 28.75 & \multicolumn{1}{c|}{6.60} & 94.58 & 98.39  \\
\multicolumn{1}{l|}{AdaNeg \cite{DBLP:conf/nips/ZhangZ24}}  & \textbf{20.40}
& \multicolumn{1}{c|}{2.79} & 94.78 & 99.26 \\
\multicolumn{1}{l|}{\cellcolor{top1}\textbf{InterNeg}}  & \cellcolor{top1}23.93 & \multicolumn{1}{c|}{\cellcolor{top1}\textbf{2.59}} & \cellcolor{top1}\textbf{95.13} & \cellcolor{top1}\textbf{99.29} \\
\bottomrule
\end{tabular}
\end{table}

\begin{table}[h!]
\small
\centering
\caption{Full results of our method with ID dataset of CIFAR10 on the OpenOOD benchmark.}
\label{tab:full_cifar10}
\begin{tabular}{l|l|c|c}
\toprule
Near / Far OOD & Datasets & FPR95 $\downarrow$ & AUROC $\uparrow$ \\
\midrule
\multirow{3}{*}{Near-OOD} & CIFAR100 \cite{Krizhevsky2009LearningML} & 37.26 & 91.81  \\
 & TIN \cite{Le2015TinyIV} & 10.60 & 98.45 \\
 & \textbf{Mean} & \textbf{23.93} & \textbf{95.13} \\
\midrule
\multirow{5}{*}{Far-OOD} & MNIST \cite{mnist} & 0.00 & 100.00 \\
 & SVHN \cite{svhn} & 0.06 & 99.97 \\
 & Texture \cite{DBLP:conf/cvpr/CimpoiMKMV14} & 0.41 & 99.57 \\
 & Places365 \cite{DBLP:journals/pami/ZhouLKO018} & 9.89 & 97.61 \\
 & \textbf{Mean} & \textbf{2.59} & \textbf{99.29} \\
\bottomrule
\end{tabular}
\end{table}

Following AdaNeg, we also evaluate our method on small-scale datasets from the OpenOOD benchmark. Specifically, we adopt CIFAR-10/100 \cite{Krizhevsky2009LearningML} as an ID dataset while utilizing CIFAR-100/10 along with TinyImageNet (TIN) \cite{Le2015TinyIV} as Near-OOD datasets. For Far-OOD datasets, we consider MNIST  \cite{mnist}, SVHN \cite{svhn}, Texture  \cite{DBLP:conf/cvpr/CimpoiMKMV14}, and Places365  \cite{DBLP:journals/pami/ZhouLKO018}. For hyperparameters, we maintain consistency with the experimental setup described in Section \ref{sec:setup}, modifying only $M$ (set to 70,000) to align with AdaNeg and ensure fair comparison. All other parameters remain unchanged. In the Near-OOD setting, our method achieves the best results on the AUROC metric.

As illustrated in Table \ref{tab:ood_results_openood_cifar100} and Table \ref{tab:ood_results_openood_cifar10}, our method still achieves state-of-the-art performance in the Far-OOD setting, even outperforming those training methods.

\subsection{Impact of ID/OOD Dataset Ordering.} 
To evaluate the robustness of our method to ID/OOD dataset ordering, we generate different dataset permutations by varying random seeds. As shown in Table \ref{tab:seed}, our method maintains stable performance with standard deviations of merely 0.05 (Four-OOD) and 0.46 (Near-OOD), demonstrating strong insensitivity to ordering effects.

\begin{table}[t]
\centering
\caption{Average results of AUROC ($\uparrow$) with five different seeds on both Four-OOD and Near-OOD benchmarks.}
\label{tab:seed}
\begin{tabular}{l|ccccc|c|c}
\toprule
\multirow{2}{*}{Benchmark} & \multicolumn{5}{c|}{Seed}             & \multirow{2}{*}{Mean} & \multirow{2}{*}{Std} \\ \cmidrule{2-6}
                                 & 0     & 1     & 2     & 3     & 4     &                       &                      \\ \midrule
Four-OOD                         & 97.43 & 97.38 & 97.38 & 97.45 & 97.51 & 97.43                 & 0.05                \\
Near-OOD                         & 82.20 & 81.56 & 81.50 & 82.29 & 82.59 & 82.03                 & 0.46                \\ \bottomrule
\end{tabular}
\end{table}

\subsection{Ablation of Different CLIP Architectures}
\label{sp:clip}

As shown in Table \ref{tab:backbone}, we evaluate multiple CLIP backbone architectures in the Four-OOD setting using ImageNet-1K as the ID dataset. The experimental results demonstrate that our method consistently outperforms all baseline approaches by a substantial margin across different backbone architectures. This robust performance advantage highlights both the effectiveness and robustness of our proposed method.

\begin{table}[h!]
\centering
\caption{OOD detection results with ID dataset of ImageNet-1k and traditional Four-OOD datasets using different CLIP backbone architectures.}
\label{tab:backbone}
\begin{adjustbox}{width=\textwidth}
\begin{tabular}{lccccccccccc}
\toprule
\multicolumn{1}{l|}{\multirow{3}{*}{Backbones}} & 
\multicolumn{1}{c|}{\multirow{3}{*}{Methods}} & \multicolumn{8}{c|}{OOD Datasets}  & \multicolumn{2}{c}{\multirow{2}{*}{Average}} \\
\multicolumn{1}{l|}{} & \multicolumn{1}{l|}{}  & \multicolumn{2}{c}{iNaturalist}     & \multicolumn{2}{c}{SUN}             & \multicolumn{2}{c}{Places}          & \multicolumn{2}{c|}{Textures}                            & \multicolumn{2}{c}{}                         \\ \cmidrule{3-12} 
\multicolumn{1}{l|}{} & \multicolumn{1}{l|}{} & AUROC$\uparrow$ & FPR95$\downarrow$ & AUROC$\uparrow$ & FPR95$\downarrow$ & AUROC$\uparrow$ & FPR95$\downarrow$ & AUROC$\uparrow$ & \multicolumn{1}{c|}{FPR95$\downarrow$} & AUROC$\uparrow$      & FPR95$\downarrow$     \\ \midrule

\multicolumn{1}{l|}{\multirow{4}{*}{ResNet50}} & \multicolumn{1}{c|}{NegLabel} & 99.24 & 2.88 & 94.54 & 26.51 & 89.72 & 42.60 & 88.40 & \multicolumn{1}{c|}{50.80} & 92.97 & 30.70 \\
 \multicolumn{1}{c|}{} & 
 \multicolumn{1}{c|}{CSP} & 99.46 & 1.95 & 95.73 & 19.05 & 90.39 & 38.58 & 92.41 & \multicolumn{1}{c|}{32.66} & 94.50 & 23.06 \\
  \multicolumn{1}{c|}{} & 
 \multicolumn{1}{c|}{AdaNeg} & \textbf{99.58} & 1.18 & 97.37 & 10.56 & \textbf{93.84} & 43.19 & 94.18 & \multicolumn{1}{c|}{35.00} & 96.24 & 22.48 \\
  \multicolumn{1}{c|}{} & 
 \multicolumn{1}{c|}{\textbf{InterNeg}} & 99.56 & \textbf{1.16} & \textbf{98.35} & \textbf{7.99} & 93.71 & \textbf{37.82} & \textbf{96.05} &   \multicolumn{1}{c|}{\textbf{21.92}} & \textbf{96.92} & \textbf{17.22} \\
\midrule

\multicolumn{1}{l|}{\multirow{4}{*}{ResNet101}} & \multicolumn{1}{c|}{NegLabel} & 99.27 & 3.11 & 94.96 & 24.55 & 89.42 & 44.82 & 87.22 & \multicolumn{1}{c|}{52.78} & 92.72 & 31.32 \\
 \multicolumn{1}{c|}{} & 
 \multicolumn{1}{c|}{CSP} & 99.47 & 2.04 & 95.71 & 19.50 & 90.27 & 39.57 & 90.59 & \multicolumn{1}{c|}{38.67} & 94.01 & 24.95 \\  \multicolumn{1}{c|}{} &
  \multicolumn{1}{c|}{AdaNeg} & \textbf{99.69} & \textbf{0.78} & 97.65 & 10.61 & \textbf{94.00} & 40.38 & 93.59 & \multicolumn{1}{c|}{39.44} & 96.23 & 22.80 \\
  \multicolumn{1}{c|}{} & 
 \multicolumn{1}{c|}{\textbf{InterNeg}} & 99.64 & 0.87 & \textbf{98.54} & \textbf{7.02} & 93.70 & \textbf{38.32} & \textbf{95.94} & \multicolumn{1}{c|}{\textbf{24.74}} & \textbf{96.96} & \textbf{17.74} \\
\midrule

\multicolumn{1}{l|}{\multirow{4}{*}{ViT-B/32}} & \multicolumn{1}{c|}{NegLabel} & 99.11 & 3.73 & 95.27 & 22.48 & 91.72 & 34.94 & 88.57 &  \multicolumn{1}{c|}{50.51} & 93.67 & 27.92 \\
 \multicolumn{1}{c|}{} & 
 \multicolumn{1}{c|}{CSP} & 99.46 & 2.37 & 96.49 & 15.01 & 92.42 & \textbf{31.47} & 93.64 &  \multicolumn{1}{c|}{25.09} & 95.50 & 18.49 \\
   \multicolumn{1}{c|}{} & 
  \multicolumn{1}{c|}{AdaNeg} & 99.67 & 0.87 & 97.74 & 9.62 & \textbf{93.98} & 36.45 & 94.58 & \multicolumn{1}{c|}{33.26} & 96.49 & 20.05 \\
  \multicolumn{1}{c|}{} & 
 \multicolumn{1}{c|}{\textbf{InterNeg}} & \textbf{99.68} & \textbf{0.70} & \textbf{98.74} & \textbf{5.79} & 93.65 & 38.05 & \textbf{96.02} & \multicolumn{1}{c|}{\textbf{23.55}} & \textbf{97.02} & \textbf{17.02} \\
\midrule

\multicolumn{1}{l|}{\multirow{4}{*}{ViT-B/16}} & \multicolumn{1}{c|}{NegLabel} & 99.49 & 1.91 & 95.49 & 20.53 & 91.64 & 35.59 & 90.22 & \multicolumn{1}{c|}{43.56} & 94.21 & 25.40 \\
\multicolumn{1}{c|}{} & 
 \multicolumn{1}{c|}{CSP}  & 99.61 & 1.54 & 96.69 & 13.82 & 92.85 & 29.69 & 93.78 & \multicolumn{1}{c|}{25.78} & 95.73 & 17.71 \\
   \multicolumn{1}{c|}{} & 
  \multicolumn{1}{c|}{AdaNeg} & 99.71 & 0.59 & 97.44 & 9.50 & 94.55 & 34.34 & 94.93 & \multicolumn{1}{c|}{31.27} & 96.66 & 18.93  \\
  \multicolumn{1}{c|}{} & 
 \multicolumn{1}{c|}{\textbf{InterNeg}} & \textbf{99.79} & \textbf{0.40} & \textbf{98.68} & \textbf{6.78} & \textbf{95.01} & \textbf{27.11} & \textbf{96.26} & \multicolumn{1}{c|}{\textbf{21.85}} & \textbf{97.43} & \textbf{14.04} \\
\midrule
\multicolumn{1}{l|}{\multirow{4}{*}{ViT-L/14}} & \multicolumn{1}{c|}{NegLabel}& 99.53 & 1.77 & 95.63 & 22.33 & 93.01 & 32.22 & 89.71 & \multicolumn{1}{c|}{42.92} & 94.47 & 24.81 \\
\multicolumn{1}{c|}{} & 
 \multicolumn{1}{c|}{CSP}  & 99.72 & 1.21 & 96.73 & 14.88 & 93.58 & 28.41 & 92.71 & \multicolumn{1}{c|}{28.16} & 95.69 & 18.17 \\
    \multicolumn{1}{c|}{} & 
  \multicolumn{1}{c|}{AdaNeg} & 99.82 & 0.26 & 97.97 & 7.94 & \textbf{95.12} & 28.67 & 94.24 & \multicolumn{1}{c|}{38.28} & 96.79 & 18.79\\
  \multicolumn{1}{c|}{} & 
 \multicolumn{1}{c|}{\textbf{InterNeg}} & \textbf{99.88} & \textbf{0.21} & \textbf{98.75} & \textbf{5.88} & 95.03 & \textbf{26.82} & \textbf{96.07} & \multicolumn{1}{c|}{\textbf{22.89}} & \textbf{97.43} & \textbf{13.95} \\
\bottomrule
\end{tabular}
\end{adjustbox}
\end{table}

\subsection{Cross-Domain Analysis}
In cross-domain settings, there is usually no ID training set provided for constructing ID image proxies. To evaluate our method in a cross-domain scenario, we conduct two additional experiments as follows:

\begin{itemize}
    \item \textbf{ImageNet ID Image Proxies:} Since ImageNet-V2 is a variant of ImageNet, we use the image proxies computed from the original ImageNet training set as a substitute.
    \item \textbf{ImageNet-V2 ID Image Proxies:} Alternatively, we randomly select a small subset of ID images (e.g., 4 images per class) from ImageNet-V2 itself to compute the ID image proxies, and use the rest for testing.
\end{itemize}

Specifically, we set ImageNet-V2 as the ID dataset and Four-OOD (iNaturalist, SUN, Places, Textures) as the OOD datasets. Table \ref{tab:cross_domain} shows the average results across the Four-OOD datasets. These results demonstrate that our method remains effective and robust in cross-domain scenarios.

\begin{table}[h]
\centering
\caption{Cross-domain OOD detection performance on ImageNet-V2 with Four-OOD datasets.}
\label{tab:cross_domain}
\begin{tabular}{llcc}
\toprule
ID Image Proxies Source & Method & AUROC $\uparrow$ & FPR95 $\downarrow$ \\
\midrule
\multirow{3}{*}{ImageNet} 
 & NegLabel & 93.08 & 29.77 \\
 & AdaNeg & 96.24 & 19.21 \\
 & \textbf{InterNeg} & \textbf{96.84} & \textbf{18.40} \\
\midrule
\multirow{3}{*}{ImageNet-V2} 
 & NegLabel & 93.90 & 27.64 \\
 & AdaNeg & 96.19 & 19.97 \\
 & \textbf{InterNeg} & \textbf{96.96} & \textbf{18.18} \\
\bottomrule
\end{tabular}
\end{table}

\subsection{ID Misclassification with Different OOD Datasets}
\label{sp::misclassification}

\begin{figure}[h]
    \centering
    \begin{subfigure}[b]{0.32\textwidth}
        \includegraphics[width=\textwidth]{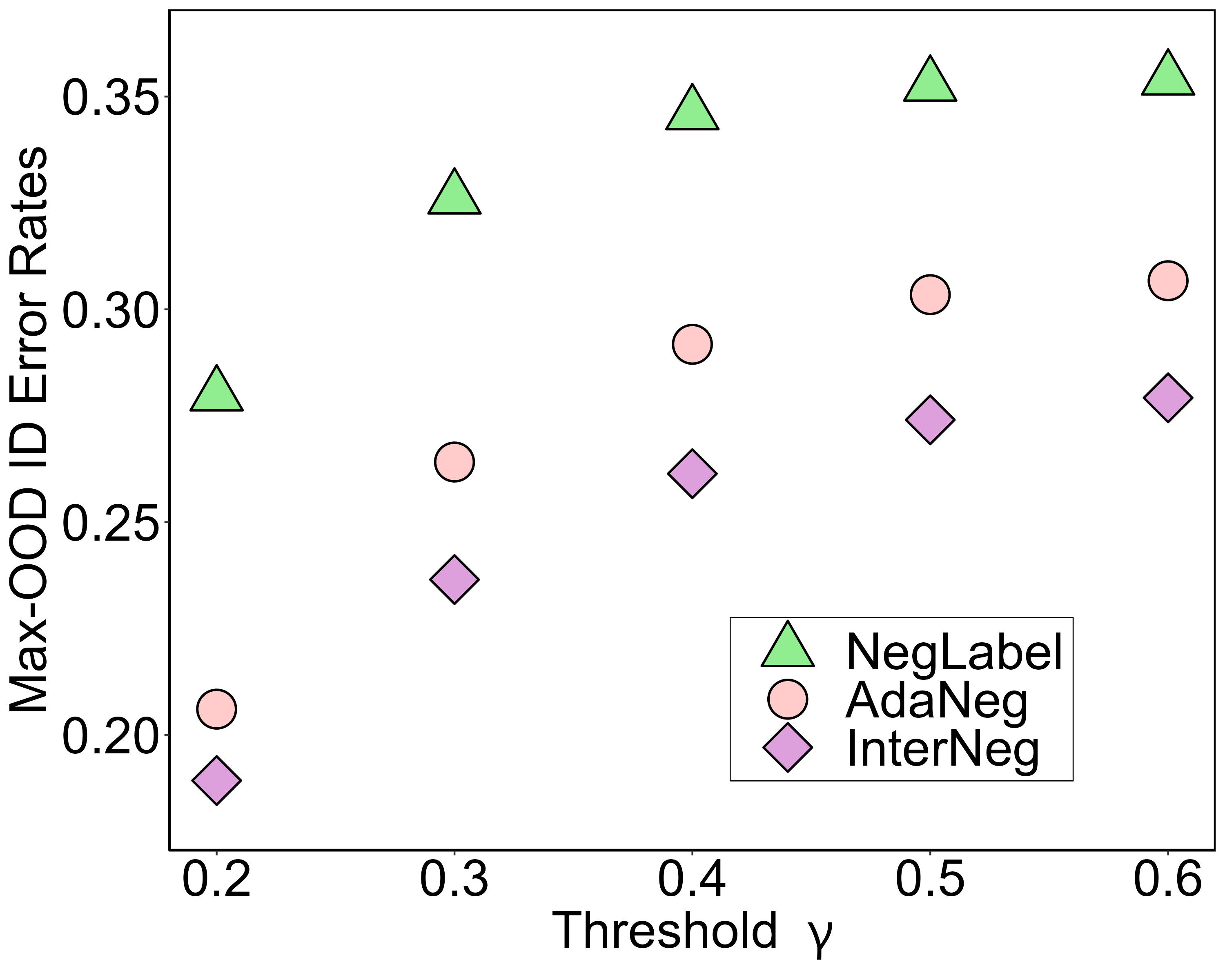}
    \end{subfigure}
    \hfill
    \begin{subfigure}[b]{0.32\textwidth}
        \includegraphics[width=\textwidth]{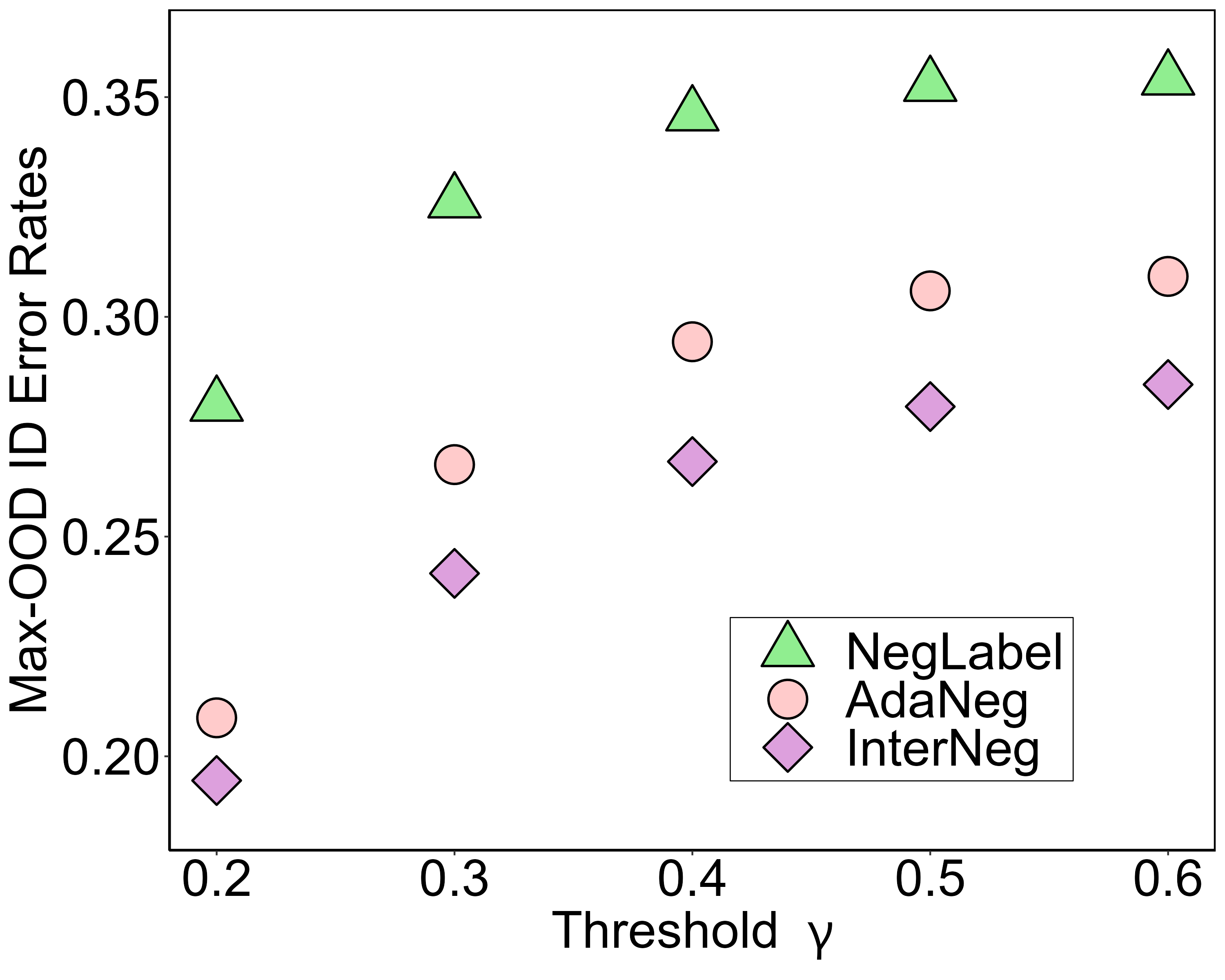}
    \end{subfigure}
    \hfill
    \begin{subfigure}[b]{0.32\textwidth}
        \includegraphics[width=\textwidth]{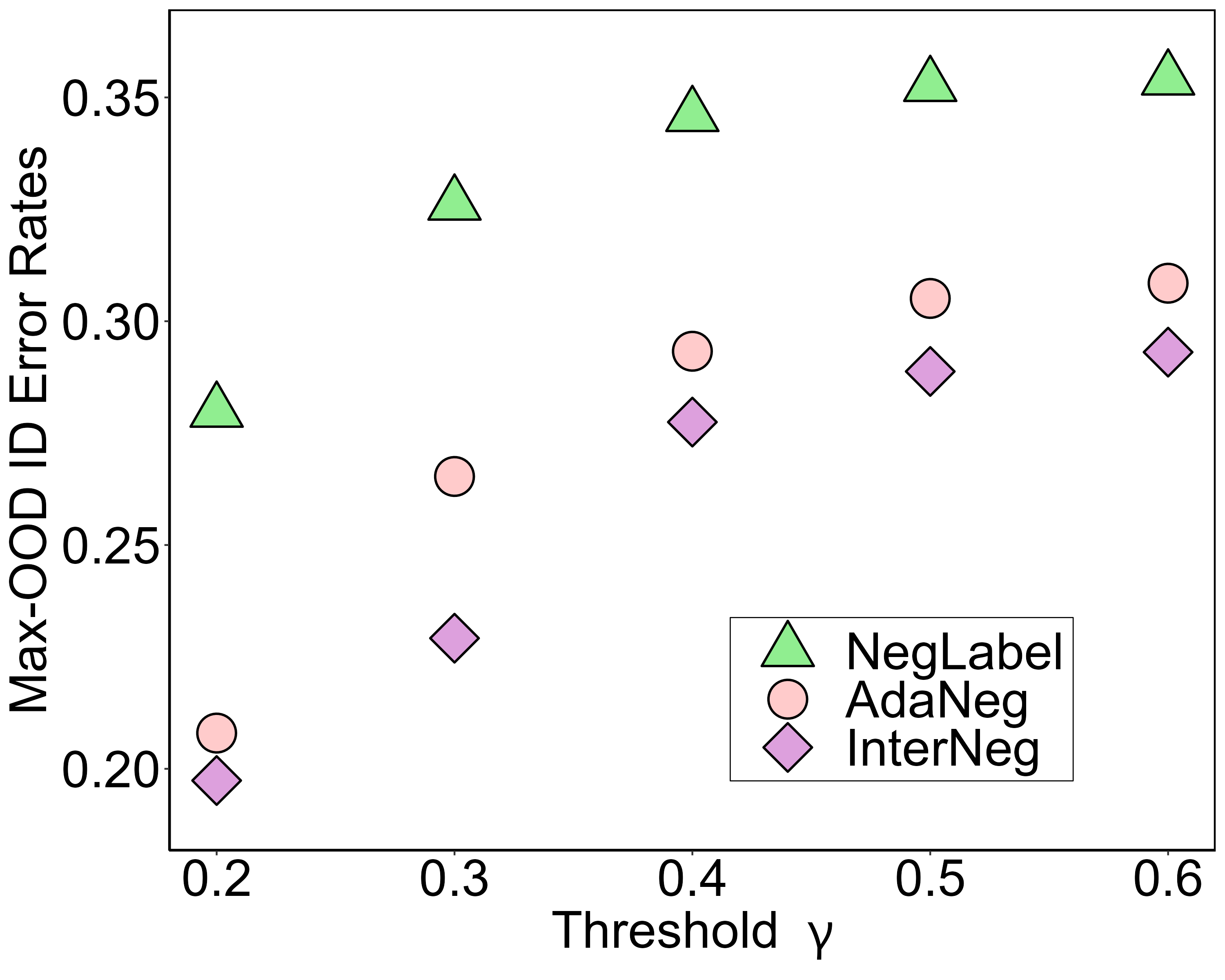}
    \end{subfigure}
    
    \vspace{0.5cm}
    
    \begin{subfigure}[b]{0.32\textwidth}
        \includegraphics[width=\textwidth]{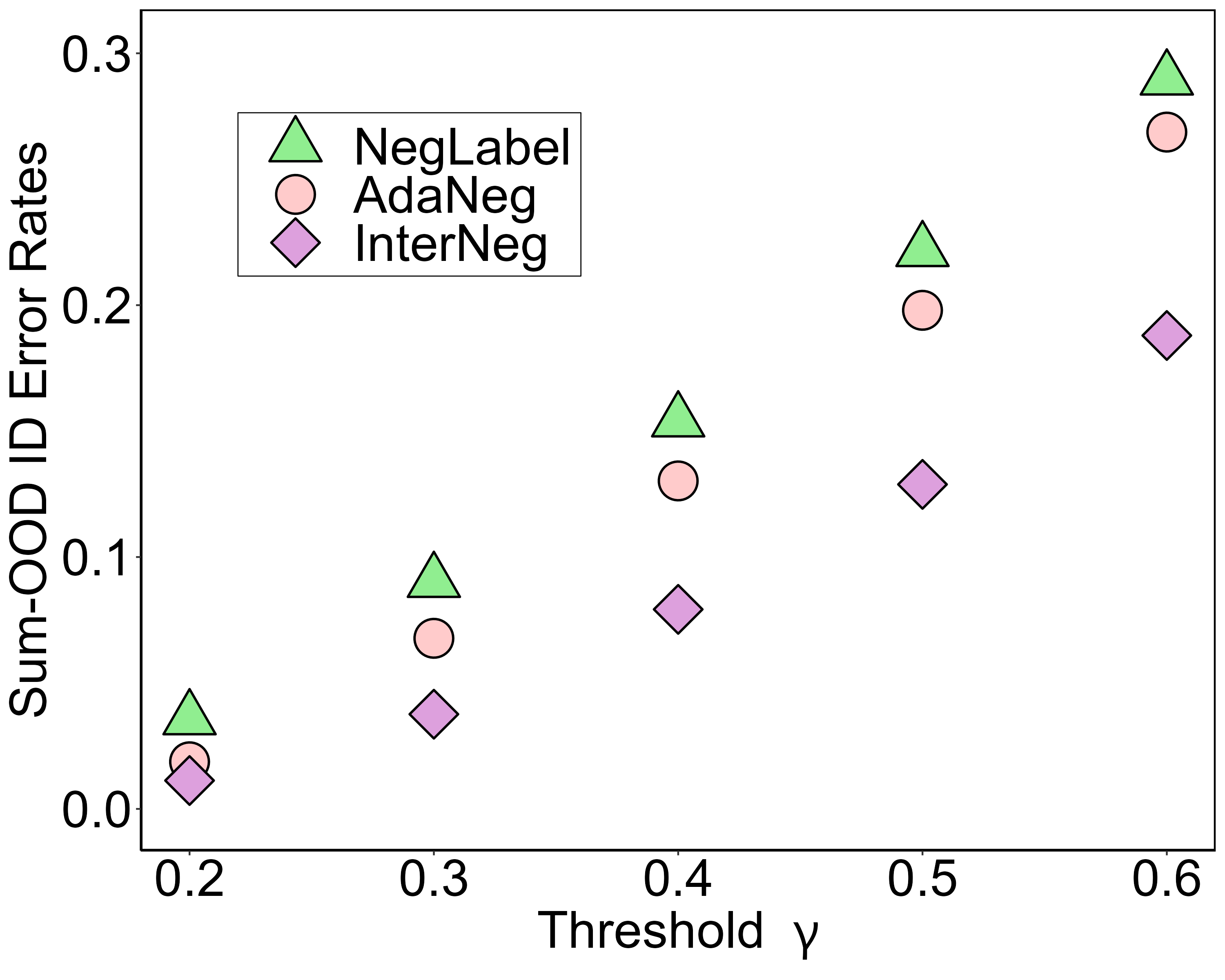}
    \end{subfigure}
    \hfill
    \begin{subfigure}[b]{0.32\textwidth}
        \includegraphics[width=\textwidth]{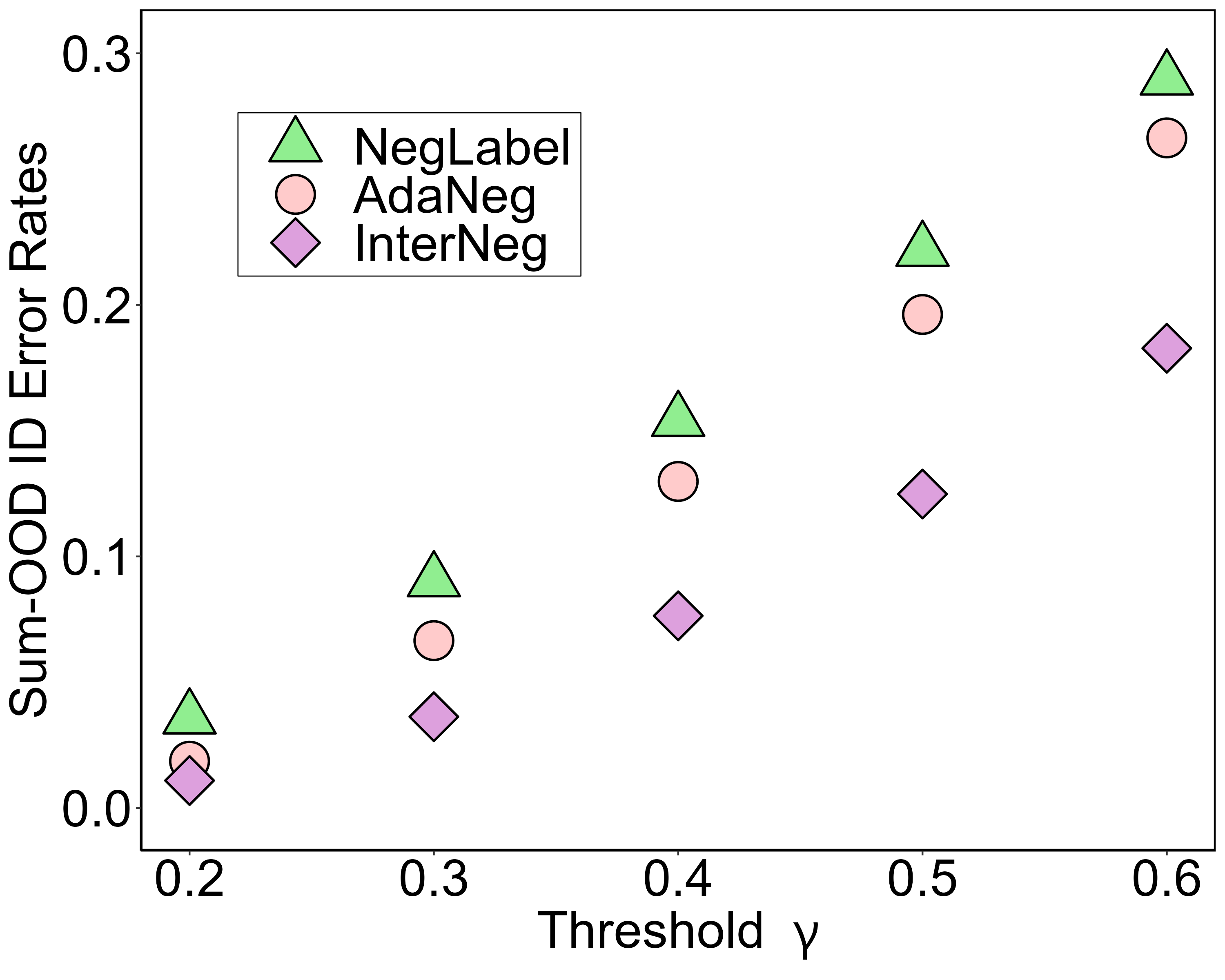}
    \end{subfigure}
    \hfill
    \begin{subfigure}[b]{0.32\textwidth}
        \includegraphics[width=\textwidth]{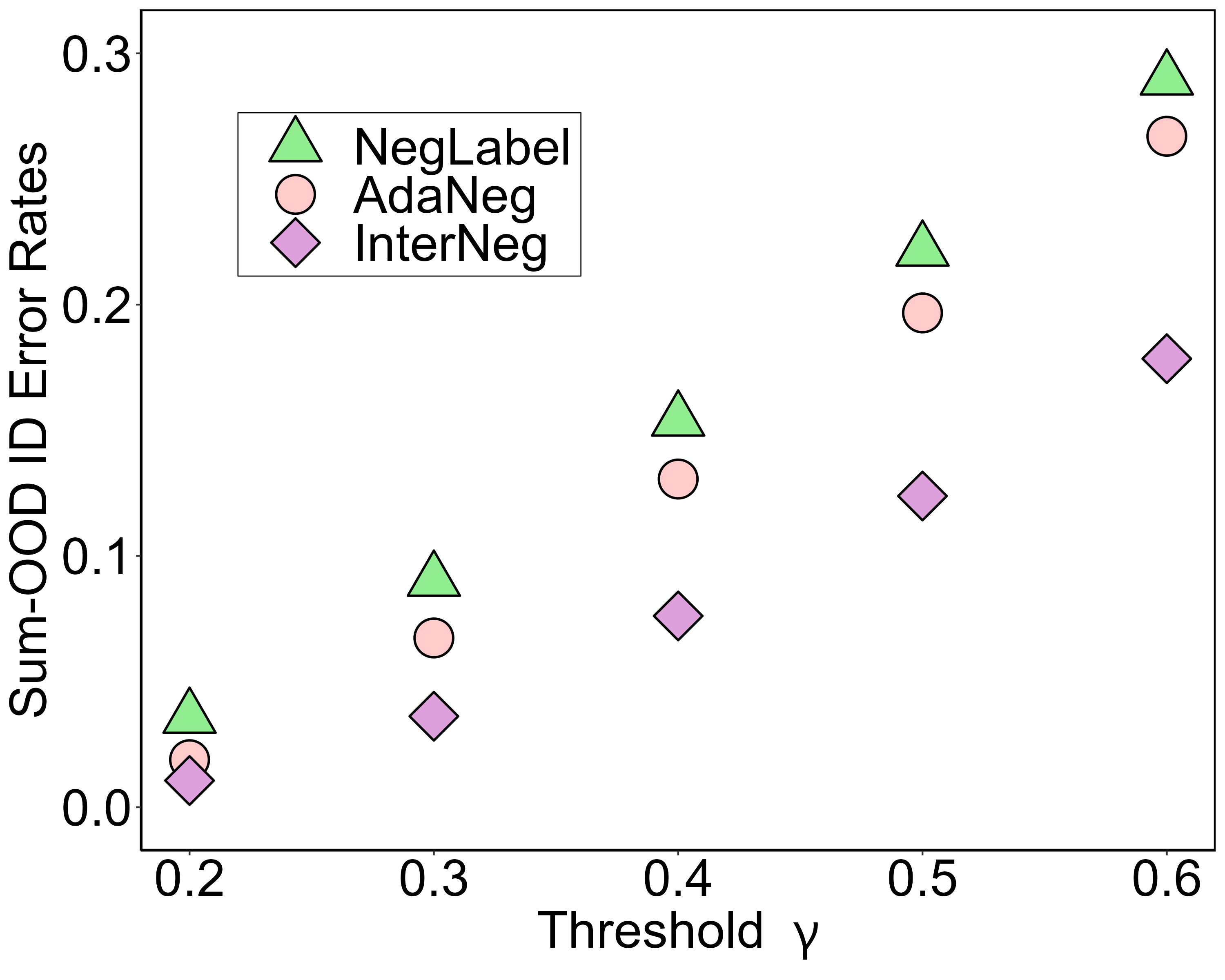}
    \end{subfigure}

    \caption{Max-OOD and Sum-OOD ID error rates on different OOD datasets. \emph{Left:} iNaturalist. \emph{Middle:} Places. \emph{Right:} Textures.}
    \label{fig:misclassification_appendix}
\end{figure}

Since both AdaNeg and our method dynamically adjust the OOD score according to the test images during inference, it is important to evaluate additional OOD datasets to better understand the phenomenon of ID misclassification. In Figure \ref{fig::misclassification}, we use SUN as the OOD dataset. Here, we extend the evaluation to other OOD datasets included in the Four-OOD benchmark. As shown in Figure \ref{fig:misclassification_appendix}, our method consistently outperforms AdaNeg across various OOD datasets, demonstrating the robustness and effectiveness of InterNeg.

\subsection{Inference Cost}
\begin{table}[H]
    \centering
    \begin{tabular}{lc|cccc}
        \toprule
        \textbf{Method} & \textbf{Mean} & \textbf{iNaturalist} & \textbf{Places} & \textbf{Textures} & \textbf{SUN} \\
        \midrule
        AdaNeg & 0.0058 & 0.0068 & 0.0046 & 0.0093 & 0.0025 \\
        \midrule
        \textbf{InterNeg}   & 0.0067 & 0.0074 & 0.0055 & 0.0113 & 0.0029 \\
        \bottomrule
    \end{tabular}
    \caption{Comparison of inference time (seconds per image).}
    \label{tab:inference_time}
\end{table}

We evaluate the inference efficiency of our approach on the traditional Four-OOD benchmark, using ImageNet-1K as the ID dataset (see Table \ref{tab:inference_time}). When measured on an NVIDIA RTX 3090 GPU, our method incurs only a \textbf{minor} computational overhead compared to AdaNeg.

\subsection{Sensitivity to Corpus Choice}

To investigate the impact of the underlying corpus, we substitute WordNet with the Common-20K and Part-of-Speech corpora. As shown in Table \ref{tab:corpus}, our method consistently outperforms the strongest baseline, AdaNeg, reducing the average FPR95 by \textbf{3.62\% and 4.76\%}, respectively. This demonstrates that our performance improvements are robust and agnostic to the choice of the source corpus.

\begin{table}[h]
    \centering
    \begin{tabular}{l|l|cc}
        \toprule
        \multirow{2}{*}{Source} & \multirow{2}{*}{Method} & \multicolumn{2}{c}{Average} \\
        \cmidrule{3-4} 
        & & AUROC $\uparrow$ & FPR95 $\downarrow$ \\
        \midrule

        \multirow{4}{*}{Common-20K} 
        & NegLabel & 90.50 & 43.02 \\
        & CSP & 92.06 & 36.56 \\
        & AdaNeg     & 93.12 & 32.39 \\
        & \textbf{InterNeg} & \textbf{94.58} & \textbf{28.77} \\
        \midrule

        \multirow{4}{*}{Part-of-Speech}
        & NegLabel & 92.71 & 32.12 \\
        & CSP & 94.21 & 24.42 \\
        & AdaNeg     & 95.07 & 23.17 \\
        & \textbf{InterNeg} & \textbf{95.98} & \textbf{18.41} \\
        \bottomrule
    \end{tabular}
    \caption{Evaluation with different corpus sources on the traditional Four-OOD benchmark, using ImageNet-1K as the ID dataset.}
    \label{tab:corpus}
\end{table}

\section{Experimental Setup}
\label{sec:setup}
\paragraph{Datasets.} Following previous work \cite{DBLP:conf/iclr/Jiang000LZ024,Fu_2025_WACV,DBLP:conf/nips/ZhangZ24,DBLP:conf/nips/Chen0X24}, we evaluate our method on the large-scale ImageNet-1K Four-OOD detection benchmark \cite{DBLP:conf/cvpr/HuangL21}. This widely-used benchmark utilizes the ImageNet-1K \cite{DBLP:conf/cvpr/DengDSLL009} dataset as ID data and iNaturalist \cite{DBLP:conf/cvpr/HornASCSSAPB18}, SUN \cite{DBLP:conf/cvpr/XiaoHEOT10}, Places \cite{DBLP:journals/pami/ZhouLKO018}, Textures \cite{DBLP:conf/cvpr/CimpoiMKMV14} as four OOD datasets, where the labels of the four OOD datasets that overlap with ImageNet-1K have been manually removed. Furthermore, we also conduct experiments on the OpenOOD benchmark \cite{DBLP:conf/nips/YangWZZDPWCLSDZ22,DBLP:journals/corr/abs-2306-09301} following \cite{DBLP:conf/nips/ZhangZ24}, This benchmark also uses ImageNet-1K as the ID dataset, while categorizing OOD data into two distinct groups based on the empirical performance of OOD detectors: Near-OOD (\textit{e.g.}, SSB-hard \cite{DBLP:conf/iclr/Vaze0VZ22}, NINCO \cite{DBLP:conf/icml/BitterwolfM023}) and Far-OOD (\textit{e.g.}, iNaturalist \cite{DBLP:conf/cvpr/HornASCSSAPB18}, Textures \cite{DBLP:conf/cvpr/CimpoiMKMV14}, OpenImage-O \cite{DBLP:conf/cvpr/Wang0F022}). Also, each OOD dataset has no classes that overlap with the ID dataset. 

\paragraph{Implementation Details.} In this paper, we implement various CLIP \cite{DBLP:conf/icml/RadfordKHRGASAM21} backbone architectures, including ResNet50, ResNet101, ViT-B/32, ViT-B/16 and ViT-L/14. Unless otherwise specified, we adopt the CLIP ViT-B/16 model as the pre-trained VLM, which consists of a visual encoder based on ViT-B/16 Vision Transformer \cite{DBLP:conf/iclr/DosovitskiyB0WZ21} and a text encoder built on Text Transformer \cite{DBLP:conf/nips/VaswaniSPUJGKP17}. For hyperparameters, we set the number $N$ of ID images per class as 16, the number $M$ of negative texts as 2000, the maximum size $K$ of extra negative text embeddings as 2000, temperature $\tau=1.0$ and threshold $\beta=0.35$ in all experiments. Following \cite{DBLP:conf/iclr/Jiang000LZ024,DBLP:conf/icml/RadfordKHRGASAM21}, we employ the text prompt template of \texttt{"The nice [class]"}. All experiments are conducted using NVIDIA RTX 3090 GPUs.

\paragraph{Evaluation Metrics.} Following common practice \cite{DBLP:conf/iclr/Jiang000LZ024,Fu_2025_WACV,DBLP:conf/nips/ZhangZ24,DBLP:conf/nips/Chen0X24}, we adopt the following metrics to evaluate the OOD detection performance: (1) AUROC, the area under the receiver operating characteristic curve; (2) FPR95, the false positive rate of the OOD data when the true positive rate of ID data is 95\%.

\section{Further Discussion}
\subsection{Comparison with ArGue and SimLabel.}

It is crucial to distinguish our approach from recent methods such as SimLabel \cite{simlabel} and ArGue \cite{ArGue}. First, we differentiate our focus on \textbf{metric consistency}—aligning the OOD detection metric with CLIP's inter-modal training objective—from SimLabel's \textbf{semantic consistency}, which focuses on aligning ID labels with semantic synonyms. Second, our method and ArGue \textbf{differ fundamentally} in their core paradigms. ArGue employs prompt tuning with negative attributes to suppress spurious visual features, aiming for robust ID classification. In contrast, InterNeg leverages inter-modal guided negative texts explicitly tailored for OOD detection.

\subsection{Clarification on Zero-Shot OOD Detection and ID Data Dependency}

\paragraph{Definition of Zero-Shot OOD Detection.}
In the context of Out-of-Distribution (OOD) detection, we align our setting with the established consensus in prior literature. For instance, NegLabel defines the zero-shot capability as predicting the correct label ``\textit{without prior training on that specific class}.'' Similarly, EOE characterizes it as operating ``\textit{without re-training on any unseen ID data}.'' Therefore, the term ``zero-shot'' typically describes the training pipeline, especially for the CLIP module. Given this constraint, it is a common practice for the baselines in this field to introduce external knowledge or additional modules. For example, ZOC /CLIPN trains a captioner/encoder to generate candidate unknown classes/negative semantics within images, respectively. To mitigate the training burden, recent advances such as NegLabel and AdaNeg turn to proxies to enhance the representations of potential OOD classes. Our method \textbf{follows this paradigm} by fixing the fundamental inconsistency hidden in the proxy generation, \textbf{without training on either ID or external data}.

\paragraph{Dependency on ID Data.}
Furthermore, assuming access to a small set of ID data is highly realistic for real-world OOD detection deployments (e.g., autonomous driving and medical diagnosis), where ID classes are inherently known and well-defined. Crucially, in our framework, we utilize ID samples \textbf{solely for proxy calculation, not model training}. As validated in Figure \ref{fig:analysis}, our approach achieves state-of-the-art performance with a minimal ID data dependency of merely 4 images per class, underscoring its practical applicability.

\end{document}